\documentclass[11pt]{article}

\usepackage[final]{acl}

\usepackage{times}
\usepackage{latexsym}
\usepackage{multirow}
\usepackage{booktabs} 
\usepackage{pifont}
\usepackage{amsmath}
\usepackage{amssymb}
\usepackage[table]{xcolor}
\usepackage{adjustbox}
\usepackage{subcaption}
\usepackage{enumitem}
\usepackage[most]{tcolorbox}
\usepackage{hyperref}

\definecolor{pink}{RGB}{255,192,203}
\definecolor{cyan}{RGB}{0,255,255}
\definecolor{green}{RGB}{0,255,0}
\definecolor{darkred}{rgb}{0.7215686274509804, 0.2235294117647059, 0.27058823529411763}
\definecolor{lightblue}{RGB}{129, 209, 241}
\definecolor{new_red}{RGB}{248, 157, 134}

\usepackage[T1]{fontenc}
\usepackage[utf8]{inputenc}

\usepackage{microtype}

\usepackage{inconsolata}

\usepackage{graphicx}

\title{QuCo-RAG: Quantifying Uncertainty from the Pre-training Corpus for Dynamic Retrieval-Augmented Generation}

\author{
 \textbf{Dehai Min\textsuperscript{1}},
 \textbf{Kailin Zhang\textsuperscript{2}},
 \textbf{Tongtong Wu\textsuperscript{3}},
 \textbf{Lu Cheng\textsuperscript{1}}
\\
\\
 \textsuperscript{1}University of Illinois at Chicago,
 \textsuperscript{2}New York University,
 \textsuperscript{3}Monash University
\\
\\
 \normalsize{
\texttt{dmin10@uic.edu, kz2739@nyu.edu, tongtong.wu@monash.edu, lucheng@uic.edu}
 }
}

\begin{document}
\maketitle

\begin{abstract}

Dynamic Retrieval-Augmented Generation adaptively determines when to retrieve during generation to mitigate hallucinations in large language models (LLMs). However, existing methods rely on model-internal signals (e.g., logits, entropy), which are fundamentally unreliable because LLMs are typically ill-calibrated and often exhibit high confidence in erroneous outputs.
We propose QuCo-RAG, which shifts from \textbf{subjective} confidence to \textbf{objective} statistics computed from pre-training data. 
Our method quantifies uncertainty through two stages: (1) before generation, we identify low-frequency entities indicating long-tail knowledge gaps; (2) during generation, we verify entity co-occurrence in the pre-training corpus, where zero co-occurrence often signals hallucination risk. Both stages leverage Infini-gram for millisecond-latency queries over 4 trillion tokens, triggering retrieval when uncertainty is high.
Experiments on multi-hop QA benchmarks show QuCo-RAG achieves EM gains of 5--12 points over state-of-the-art baselines with OLMo-2 models, and transfers effectively to models with undisclosed pre-training data (Llama-3, Qwen2.5, GPT-4.1/5-chat), improving EM by up to 14 points. Generalization to long-form generation and biomedical QA further validates the robustness of our paradigm.
These results establish corpus-grounded verification as a principled, practically model-agnostic paradigm for dynamic RAG\footnote{Our code is publicly available at \url{https://github.com/ZhishanQ/QuCo-RAG}.}.

% https://github.com/ZhishanQ/QuCo-RAG

\end{abstract}

\section{Introduction}

Retrieval-Augmented Generation (RAG)~\cite{lewis2020retrieval,gao2023retrieval} mitigates LLM hallucinations by grounding generation in external evidence. Early RAG systems employ static strategies with a single retrieval step before generation~\cite{karpukhin-etal-2020-dense, shi2024replug, min-etal-2025-unihgkr}, but fall short for complex multi-step tasks where information needs emerge dynamically during generation~\cite{10.1145/3735127,wang2025llms,wang-etal-2023-self-knowledge}. This has driven the emergence of Dynamic RAG methods that adaptively determine when and what to retrieve based on the generation process~\cite{jiang-etal-2023-active,asai2024selfrag}.

\begin{figure}[t]
    \centering
    \includegraphics[width=0.98\linewidth]{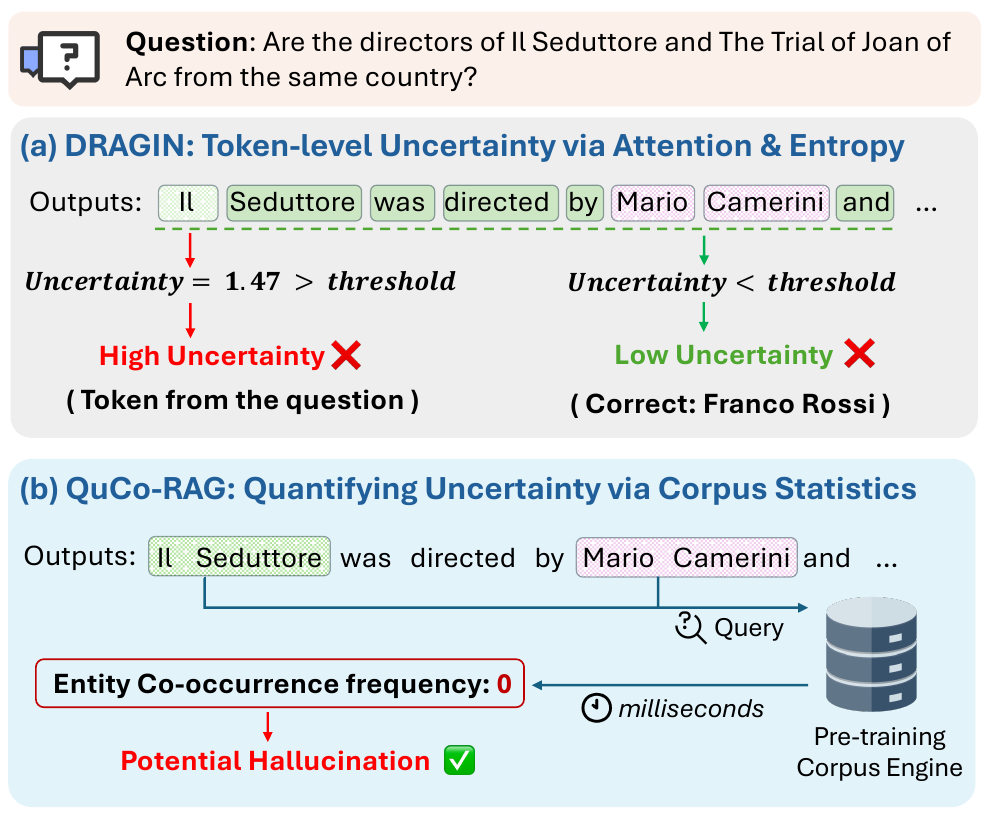}
    \caption{Comparison of retrieval triggering mechanisms. (a) DRAGIN relies on model-internal signals, incorrectly assigning high uncertainty to ``Il'' (a token from the question) while showing low uncertainty on the hallucinated director name. (b) QuCo-RAG correctly detects the hallucination through zero entity co-occurrence in the pre-training corpus.}
    \label{fig:figure_1}
    \vspace{-0.65cm}
\end{figure}

Current dynamic RAG methods predominantly rely on quantifying uncertainty through model-internal signals such as token probability~\cite{jiang-etal-2023-active} or entropy~\cite{su-etal-2024-dragin,li2025modeling} to determine when to retrieve. However, these methods assume internal signals reliably indicate generation correctness, an assumption that is fundamentally flawed~\cite{li2024survey}. As illustrated in Figure~\ref{fig:figure_1}(a), the notable work DRAGIN \cite{su-etal-2024-dragin} exhibits low uncertainty when generating the incorrect director name ``Mario Camerini'', yet assigns high uncertainty to ``Il'', a token from the question.
This failure reflects a well-documented problem: LLMs are poorly calibrated~\cite{guo2017calibration,kadavath2022language,achiam2023gpt}---their confidence scores fail to correlate with actual prediction accuracy. This miscalibration leads to ``confident hallucinations,'' where models produce incorrect content with high confidence~\cite{tian-etal-2023-just}. Furthermore, post-training techniques such as SFT~\cite{dong2024abilities} and RLHF~\cite{ouyang2022training} often exacerbate this by encouraging decisive answers. More fundamentally, recent theoretical work~\cite{kalai2024calibrated} further shows that for rarely-seen facts, even perfectly calibrated models must hallucinate to maintain statistical consistency.
% guo2025deepseek

To bypass the limitations, we propose \textbf{QuCo-RAG}, a framework that determines when to retrieve by \textbf{Qu}antifying uncertainty via pre-training \textbf{Co}rpus statistics, shifting from subjective internal confidence to objective external evidence. Our key insight is that an LLM's factual knowledge is fundamentally shaped by its pre-training corpus~\cite{balepur2025reverse}: 
low-frequency entities correspond to long-tail knowledge that models struggle to memorize reliably, while zero co-occurrence between entity pairs indicates the model has no evidential basis for claims relating them. Based on this insight, QuCo-RAG operates through two-stage detection: \textbf{(1) Pre-Generation Knowledge Assessment:} We query entity frequencies in the pre-training corpus, triggering retrieval when entities are low-frequency (long-tail knowledge risks).
\textbf{(2) Runtime Claim Verification:} We extract knowledge triplets from each generated sentence and verify entity co-occurrence; zero co-occurrence triggers retrieval and regeneration.
Both stages leverage Infini-gram~\cite{liu2024infinigram} for millisecond-latency queries over trillion-token corpora.

To validate our approach, we first evaluate QuCo-RAG on multi-hop QA benchmarks using the OLMo-2 model family (7B, 13B, 32B)~\cite{olmo20242}, which provides full access to its 4-trillion token pre-training corpus for precise statistical verification. Results show QuCo-RAG achieves 5--12 point improvements on Exact Match (EM) over state-of-the-art baselines across all model scales, while maintaining competitive efficiency.

Beyond this matched-corpus setting, we demonstrate QuCo-RAG's broad applicability through two additional dimensions of evaluation. First, for \textbf{cross-model transferability}, we show that corpus statistics computed from OLMo-2's pre-training corpus serve as effective proxies for models with undisclosed training data. Leveraging the substantial overlap of web-scale pre-training corpora, QuCo-RAG yields up to 14 EM improvements on Llama-3, Qwen2.5, and GPT-4.1/5-chat series. Second, for \textbf{task and domain generalization}, we evaluate on ASQA~\citep{stelmakh-etal-2022-asqa}, a long-form generation benchmark, and PubMedQA~\cite{jin2019pubmedqa}, a biomedical QA benchmark. QuCo-RAG achieves the best performance on both while internal-signal methods show limitations in either efficiency or effectiveness, demonstrating that our framework generalizes robustly without task- or domain-specific tuning.

\section{Related Work}

\begin{figure*}[th!]
    \centering
    \includegraphics[width=0.9\textwidth]{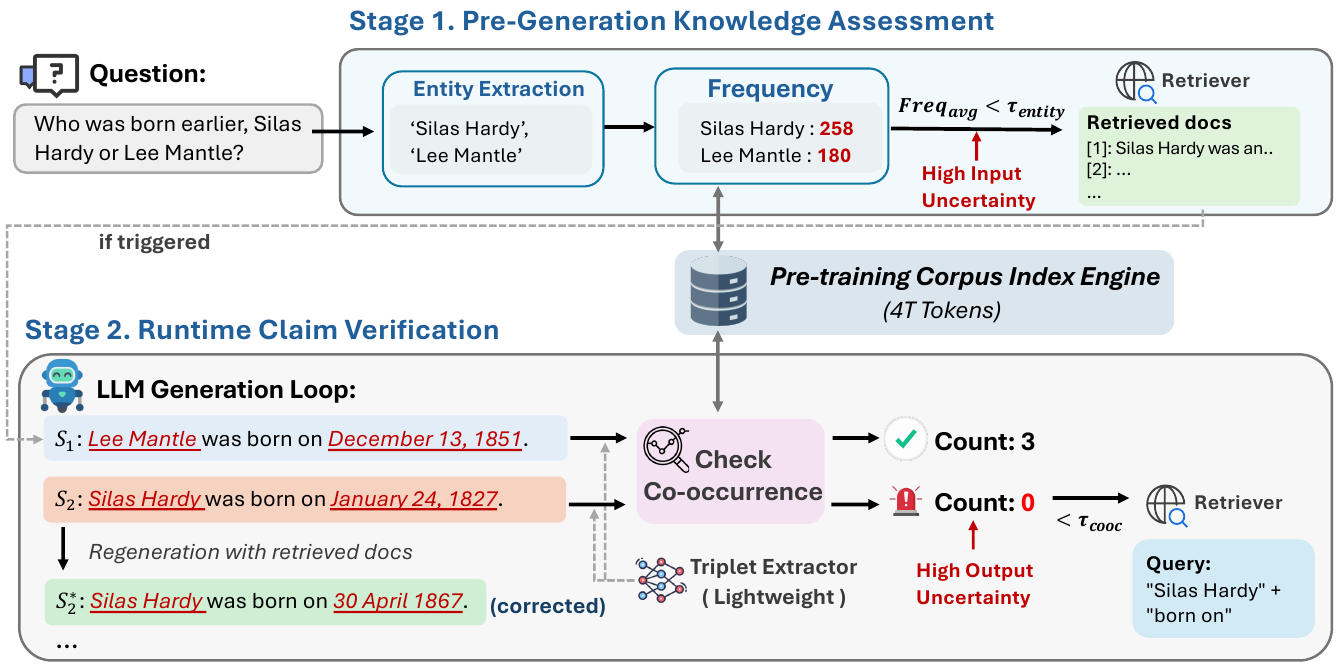}
    \vspace{-0.2cm}
    \caption{Overview of QuCo-RAG Framework. 
    % Stage 1 extracts entities from the question and queries their frequencies in the pre-training corpus, triggering retrieval when entity frequency falls below a threshold. Stage 2 verifies each generated sentence by checking entity co-occurrence; zero co-occurrence triggers retrieval and regeneration.
    }
    \label{fig:method_overview}
    \vspace{-0.4cm}
\end{figure*}

\paragraph{Dynamic Retrieval-Augmented LLM}
Dynamic RAG methods have evolved to address the limitations of static retrieval approaches by adaptively determining when and what to retrieve during generation~\cite{xu2024activerag,yu2024auto,yang2025knowing}. FLARE \cite{jiang-etal-2023-active} pioneered this direction by triggering retrieval when encountering low-probability tokens.
Self-RAG \cite{asai2024selfrag} extended this paradigm by training models to generate special reflection tokens that assess retrieval necessity and response quality, though requiring additional fine-tuning. More recent approaches~\cite{ma2025estimating} construct more sophisticated uncertainty metrics: DRAGIN~\cite{su-etal-2024-dragin} integrates multiple model-internal signals including entropy and attention weights, ETC~\cite{li2025modeling} considers first- and second-order entropy differences to capture uncertainty trends, and SeaKR~\cite{yao2025seakr} extracts self-aware uncertainty from LLMs' internal FFN states.
However, these methods all rely on model-internal signals, which may not reliably indicate correctness.

\paragraph{Reusing LLM Pre-Training Data at Inference Time}
Recent work explores unlocking additional value from pre-training corpora at inference time. \citet{fang2025reusing} showed that retrieving from the model's own pre-training data yields performance gains equivalent to a $5\times$ increase in pre-training compute. Efficient infrastructure has emerged to support trillion-scale corpus access. Infini-gram~\cite{liu2024infinigram} provides millisecond-latency $n$-gram counting via suffix arrays, while Infini-gram mini~\cite{xu-etal-2025-infini} reduces index size to 44\% of the corpus via FM-index~\cite{ferragina2000opportunistic}. OLMoTrace~\cite{liu2025olmotrace} enables real-time tracing of LLM output back to verbatim matches in training documents.
Our work leverages this infrastructure for a distinct purpose: 
using pre-training corpus statistics to \emph{quantify uncertainty and trigger retrieval}, enabling reliable hallucination detection and mitigation.

% using corpus statistics as \emph{retrieval triggers} rather than retrieval content, enabling reliable hallucination detection.
\section{Methodology}

\subsection{Problem Formulation}

We formalize the dynamic RAG problem as follows. Let $\mathcal{M}$ denote an LLM, $\mathcal{C}$ represent an external knowledge base for retrieval (e.g., Wikipedia), and $\mathcal{P}$ denote the pre-training corpus used to train $\mathcal{M}$. Given an input question $Q$, the model generates a response $y = (s_1, s_2, \ldots, s_N)$, where $s_i$ is the $i$-th generated sentence.
A dynamic RAG system makes two critical decisions during generation:

\noindent\textbf{(1) When to retrieve.} At each step $i$, determine whether to trigger retrieval:
\begin{equation}
\delta_i = f_{\text{trigger}}(Q, s_{\leq i}; \Theta) \in \{0, 1\},
\end{equation}
where $\Theta$ denotes the source of uncertainty signals. Unlike prior methods that rely on internal model states (i.e., $\Theta = \mathcal{M}$), we ground the decision in pre-training corpus statistics (i.e., $\Theta = \mathcal{P}$).

\noindent\textbf{(2) What to retrieve.} When $\delta_i = 1$, construct a query $q_i = f_{\text{query}}(Q, s_{\leq i})$ and retrieve related documents $\mathcal{D}_i = \text{Retrieve}(q_i, \mathcal{C})$, where $f_{\text{query}}$ is the query formulation function.

\paragraph{Binary Nature of Retrieval Decisions.}
Note that the retrieval decision $\delta_i \in \{0, 1\}$ is inherently binary: the system either retrieves or not.
% proceeds without retrieval.
This observation motivates our design: rather than estimating \textit{continuous confidence scores} from model-internal signals to infer uncertainty, whose thresholds lack clear semantic grounding, we directly leverage \textit{discrete corpus statistics} to determine whether the model faces high uncertainty (retrieve) or low uncertainty (proceed without retrieval).
Specifically, we consider two high-uncertainty scenarios:
\textbf{(1) Input uncertainty}: the question contains entities rarely seen during pre-training, indicating insufficient knowledge coverage;
\textbf{(2) Output uncertainty}: the generated claim relates entities that never co-occur in the corpus, indicating lack of evidential support.
Both signals are grounded in corpus statistics, as illustrated in Figure~\ref{fig:method_overview}.
% operationalized through two stages 

\subsection{Pre-Generation Knowledge Assessment}

To quantify input uncertainty, we employ a pre-check mechanism before generation begins. 
We first use a lightweight entity extractor to identify a set of key entities $\mathcal{E}_Q = \{e_1, e_2, \ldots, e_m\}$ from the input question $Q$. For each entity $e \in \mathcal{E}_Q$, we query its frequency in the pre-training corpus $\mathcal{P}$, denoted as $\text{freq}(e; \mathcal{P})$.
We posit that entities with low frequency in $\mathcal{P}$ represent long-tail knowledge risks, where the model is likely to hallucinate. Retrieval is triggered if the average entity frequency falls below a predefined threshold:
\begin{equation}
\delta_{\text{pre}} = \mathbb{I}\left(
% \frac{1}{|\mathcal{E}_Q|}\sum
Avg_{e \in \mathcal{E}_Q}\text{freq}(e; \mathcal{P})
< \tau_{\text{entity}}
\right).
\end{equation}
We set $\tau_{\text{entity}} = 10^3$ as the default threshold; results remain stable across a wide range ($10^3$ to $10^7$) as shown in Appendix~\ref{sec:threshold_sensitivity}. If $\delta_{\text{pre}} = 1$, we use the original question $Q$ as the search query to retrieve relevant documents $\mathcal{D}_0$, which are prepended to the model's context before generation starts.

\subsection{Runtime Claim Verification}
\label{sec:runtime_verification}

To quantify output uncertainty, QuCo-RAG continuously monitors each generated sentence $s_i$ by verifying whether the claimed facts have evidential support in the pre-training corpus. For a generated sentence $s_i$, we extract a set of knowledge triplets $\mathcal{T} = \{(h, r, t)\}$, where $h$, $r$, $t$ represent the head entity, relation, and tail entity, respectively. We quantify the evidential support for each triplet by computing the co-occurrence frequency of the head and tail entities within a defined window $\omega$ (e.g., a document or paragraph) in $\mathcal{P}$:
\begin{equation}
% \text{cooc}(h, t; \mathcal{P}) = \text{Count}(h \in \omega \land t \in \omega \mid \omega \in \mathcal{P})
\text{cooc}(h, t; \mathcal{P}) = |\{\omega \in \mathcal{P} : h \in \omega \land t \in \omega\}|.
\end{equation}
We compute $\text{cooc}(h, t)$ rather than $\text{cooc}(h, r, t)$ because relational predicates exhibit high lexical variability (e.g., ``employed by'' vs. ``worked at''), while named entities are more lexically stable~\cite{galarraga2014canonicalizing}.
Retrieval is triggered if the co-occurrence count falls below a threshold $\tau_{\text{cooc}}$ (default set to 1):
\begin{equation}
% \delta_i = \mathbb{I}\left( \exists (h, r, t) \in \mathcal{T} : \text{cooc}(h, t; \mathcal{P}) < \tau_{\text{cooc}} \right).
\delta_i = \mathbb{I}\left( \min_{(h,r,t) \in \mathcal{T}} \text{cooc}(h, t; \mathcal{P}) < \tau_{\text{cooc}} \right).
\end{equation}
We use minimum rather than average here because a single unsupported claim suffices to indicate hallucination risk, whereas Stage~1 uses average to capture overall knowledge coverage (see Table~\ref{tab:aggregation} for an empirical comparison).
The rationale for $\tau_{\text{cooc}} = 1$ is intuitive: if two entities never co-occur in the pre-training corpus, the generated claim lacks evidential support and likely constitutes a hallucination~\cite{mallen-etal-2023-trust,kandpal2023large}. Notably, co-occurrence evidence is \emph{asymmetric}: while $\text{cooc}(h, t; \mathcal{P}) > 0$ does not guarantee correctness (entities may co-occur with different relations or in unrelated contexts), $\text{cooc}(h, t) = 0$ strongly indicates hallucination risk~\cite{gao-etal-2023-enabling,ravichander-etal-2025-halogen}. We quantify the impact of this asymmetry in Appendix~\ref{sec:cooc_analysis}.
When retrieval is triggered ($\delta_i = 1$), we construct a \textit{Semantic-Oriented Query} using the head entity and relation ($q = h \oplus r$) to retrieve supporting documents and regenerate the sentence.

\subsection{Implementation Details}

\paragraph{Corpus Statistics via Infini-gram.}
We leverage Infini-gram~\cite{liu2024infinigram}, a suffix array-based engine that supports millisecond-latency queries over trillion-token corpora, enabling real-time computation during generation. Local deployment resource requirements are detailed 
in Appendix~\ref{app:prompts}.

\paragraph{Lightweight Triplet Extraction.}
To minimize overhead while ensuring extraction quality, we distill a specialized 0.5B model from GPT-4o-mini~\cite{hurst2024gpt}. Specifically, we construct 40K annotated examples using in-context learning, then perform full-parameter supervised fine-tuning on Qwen2.5-0.5B-Instruct~\cite{qwen2.5}. Representative training examples are provided in Appendix~\ref{app:triplet_examples}. Intrinsic evaluation (89.9\% entity-level F1) and an end-to-end ablation confirming that the 0.5B fine-tuned extractor matches the GPT-4o-mini teacher are reported in Appendix~\ref{app:extractor_eval}.

\section{Experimental Setup}

\begin{table*}[th!]
    \centering
    \caption{Performance comparison on multi-hop QA benchmarks across OLMo-2 model scales. \textbf{Bold}: best; \underline{underline}: second-best. \textit{Improv.}: absolute gain over best baseline.  2Wiki: 2WikiMultihopQA.}
    
    \label{tab:main_results}
    \begin{adjustbox}{max width=0.88\textwidth}
    \begin{tabular}{l cccc cccc cccc}
        \toprule
        & \multicolumn{4}{c}{\textbf{OLMo-2-7B}} & \multicolumn{4}{c}{\textbf{OLMo-2-13B}} & \multicolumn{4}{c}{\textbf{OLMo-2-32B}} \\
        \cmidrule(lr){2-5} \cmidrule(lr){6-9} \cmidrule(lr){10-13}
        & \multicolumn{2}{c}{2Wiki} & \multicolumn{2}{c}{HotpotQA} & \multicolumn{2}{c}{2Wiki} & \multicolumn{2}{c}{HotpotQA} & \multicolumn{2}{c}{2Wiki} & \multicolumn{2}{c}{HotpotQA} \\
        \cmidrule(lr){2-3} \cmidrule(lr){4-5} \cmidrule(lr){6-7} \cmidrule(lr){8-9} \cmidrule(lr){10-11} \cmidrule(lr){12-13}
        Method & EM & F1 & EM & F1 & EM & F1 & EM & F1 & EM & F1 & EM & F1 \\
        \midrule
        Wo-RAG & 20.1 & 26.4 & 22.6 & 31.6 & 28.5 & 34.5 & 24.4 & 33.6 & 33.3 & 40.3 & 22.0 & 31.3 \\
        SR-RAG & 23.7 & 30.7 & \underline{29.7} & \underline{40.7} & 28.9 & 35.7 & \underline{29.7} & \underline{39.5} & \underline{37.4} & \underline{46.5} & 29.5 & 40.4 \\
        FS-RAG & 21.1 & 28.3 & 14.5 & 20.7 & 28.8 & 35.1 & 14.6 & 21.9 & 34.6 & 41.0 & 13.9 & 19.5 \\
        FLARE & 22.9 & 28.9 & 20.3 & 28.4 & 26.2 & 31.5 & 15.3 & 21.9 & 32.0 & 39.3 & 28.3 & 39.8 \\
        DRAGIN & 22.8 & 29.0 & 17.5 & 24.7 & 28.5 & 33.9 & 19.5 & 27.6 & 33.3 & 40.2 & 17.7 & 24.3 \\
        ETC & 23.4 & 29.8 & 25.1 & 34.7 & \underline{29.7} & \underline{35.9} & 29.3 & \underline{39.5} & 36.0 & 43.6 & \underline{30.8} & \underline{42.2} \\
        SeaKR & \underline{25.3} & \underline{32.7} & 24.8 & 35.0 & 29.6 & 34.6 & 26.2 & 37.3 & 30.2 & 38.2 & 28.7 & 40.4 \\
        \midrule
        \rowcolor{lightblue!20}
        QuCo-RAG & \textbf{32.7} & \textbf{41.1} & \textbf{35.3} & \textbf{46.1} & \textbf{41.7} & \textbf{49.1} & \textbf{35.0} & \textbf{46.8} & \textbf{46.8} & \textbf{56.2} & \textbf{41.6} & \textbf{54.2} \\
        \rowcolor{lightblue!20}
        \textit{Improv.} & \textcolor{darkred}{\textbf{+7.4}} & \textcolor{darkred}{\textbf{+8.4}} & \textcolor{darkred}{\textbf{+5.6}} & \textcolor{darkred}{\textbf{+5.4}} & \textcolor{darkred}{\textbf{+12.0}} & \textcolor{darkred}{\textbf{+13.2}} & \textcolor{darkred}{\textbf{+5.3}} & \textcolor{darkred}{\textbf{+7.3}} & \textcolor{darkred}{\textbf{+9.4}} & \textcolor{darkred}{\textbf{+9.7}} & \textcolor{darkred}{\textbf{+10.8}} & \textcolor{darkred}{\textbf{+12.0}} \\
        \bottomrule
    \end{tabular}
    \end{adjustbox}
    \vspace{-0.3cm}
\end{table*}

\subsection{Datasets and Implementation}

% We evaluate on two widely adopted knowledge-intensive multi-hop QA benchmarks: 2WikiMultihopQA~\cite{ho-etal-2020-constructing} and HotpotQA~\cite{yang2018hotpotqa}, sampling the first 1,000 validation examples from each as our test sets following~\citet{su-etal-2024-dragin}. We report Exact Match (EM) and token-level F1 score as evaluation metrics.

We evaluate on two widely adopted knowledge-intensive multi-hop QA benchmarks: 2WikiMultihopQA~\cite{ho-etal-2020-constructing} and HotpotQA~\cite{yang2018hotpotqa}. 
Following~\citet{su-etal-2024-dragin}, we sample the first 1,000 validation examples from each as our test sets and report Exact Match (EM) and token-level F1 score as evaluation metrics, which are well-suited for these benchmarks as answers are short-form entities that can be reliably extracted and matched. Prior work~\cite{li2025modeling} has shown that EM/F1 conclusions align with LLM-as-judge~\cite{li2025generation} evaluations on these datasets.
For retrieval, we employ BM25~\cite{robertson2009probabilistic} over the Wikipedia dump from~\citet{karpukhin-etal-2020-dense} as our external corpus $\mathcal{C}$, retrieving top-3 documents per query. We also verify robustness with dense retrievers in Appendix~\ref{app:retriever_robustness}.
In our experiments, we query entity frequencies and co-occurrences via the Infini-gram API\footnote{API Endpoint Documentation: \url{https://infini-gram.readthedocs.io/en/latest/api.html}. The Infini-gram index supports local deployment for offline environments, requiring primarily CPU and disk storage rather than GPU resources.}, which hosts the full OLMo-2 pre-training corpus index. We set the co-occurrence window size to 1,000 tokens, roughly matching passage-level context length; sensitivity analysis across $\omega = 50$ to $2{,}000$ is provided in Appendix~\ref{sec:threshold_sensitivity}.
More detailed LLM generation settings and the full prompt template are provided in Appendix~\ref{app:prompts}.
All experiments are conducted on NVIDIA H200 GPUs (141GB HBM3e).

\subsection{Baselines}

\noindent\textbf{No Retrieval:}
\textbf{Wo-RAG} generates answers directly without any external retrieval, serving as the lower bound to measure RAG benefits.

\noindent\textbf{Static Retrieval:}
Single-Round RAG (\textbf{SR-RAG}): performs one-time retrieval using the input question before generation begins.
Fixed-Sentence RAG (\textbf{FS-RAG})~\cite{trivedi2023interleaving}, also known as IRCoT, triggers retrieval after every generated sentence, using the last sentence as the query.

\noindent\textbf{Dynamic Retrieval:}
\textbf{FLARE}~\cite{jiang-etal-2023-active} triggers retrieval on low-probability tokens.
\textbf{DRAGIN}~\cite{su-etal-2024-dragin} combines entropy, attention, and semantic signals.
\textbf{ETC}~\cite{li2025modeling} models first- and second-order entropy differences.
\textbf{SeaKR}~\cite{yao2025seakr} leverages internal FFN states for uncertainty estimation.
All baseline results are reproduced using their released code.

\subsection{Models}

\paragraph{Primary Models (Matched Corpus).}

We use the OLMo-2-Instruct family~\cite{olmo20242} (7B, 13B, and 32B) as our primary evaluation targets. OLMo-2 achieves performance competitive with mainstream models like Qwen2.5 while providing publicly available training data, code, and recipes. The pre-training corpus\footnote{\url{https://huggingface.co/datasets/allenai/olmo-mix-1124}} comprises about 4 trillion tokens from diverse sources. This transparency enables precise computation of entity frequencies and co-occurrence statistics, making OLMo-2 ideal for validating our method.

\paragraph{Transferability Models (Proxy Corpus).}
A key advantage of QuCo-RAG is its applicability to LLMs with undisclosed pre-training data. Given that web-scale pre-training corpora share substantial overlap~\cite{soldaini2024dolma}, statistics derived from a transparent and comprehensive corpus can serve as effective proxies for other models. 
We demonstrate this by using the OLMo-2 corpus as a proxy for Llama-3-8B-Instruct~\cite{grattafiori2024llama}, Qwen2.5-32B-Instruct~\cite{qwen2.5}, and proprietary models (GPT-4.1~\cite{OpenAI:GPT-4_1}, GPT-5-chat~\cite{OpenAI:GPT-5}). For GPT models, we additionally compare against their built-in agentic web search, where the model autonomously invokes web search via the Responses API.

\section{Experimental Results}
We design experiments to answer three core research questions:
\begin{itemize}[leftmargin=*, itemsep=0pt, topsep=2pt]
    \item \textbf{RQ1:} How does corpus-based uncertainty compare to model-internal signals? (\S\ref{sec:main_results})
    \item \textbf{RQ2:} How well does QuCo-RAG transfer to models with undisclosed training data? (\S\ref{sec:transferability})
    \item \textbf{RQ3:} What is the efficiency-performance trade-off of QuCo-RAG? (\S\ref{sec:efficiency})
\end{itemize}

\subsection{Main Results (RQ1)}
\label{sec:main_results}

\begin{figure*}[h!]
    \centering
    \includegraphics[width=\linewidth]{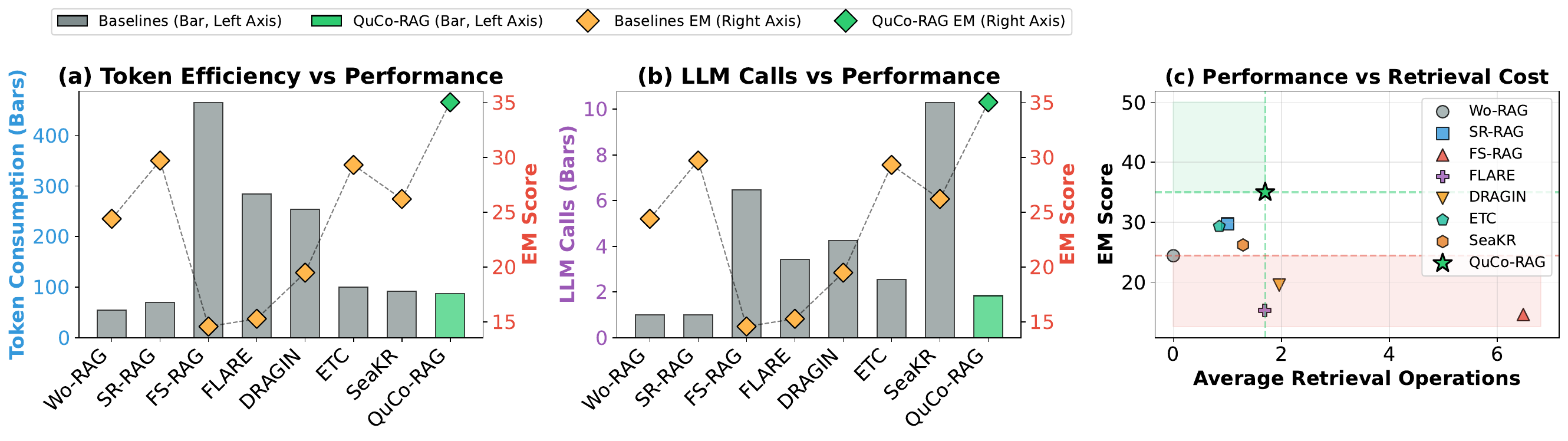}
    \vspace{-0.6cm}
    \caption{Efficiency-performance trade-off analysis on HotpotQA with OLMo-2-13B-Instruct. (a) EM score versus Token consumption. (b) EM score versus LLM calls. (c) Performance versus Retrieval frequency. QuCo-RAG achieves the highest EM with moderate token usage and LLM calls. }
    \label{fig:efficiency}
    \vspace{-4mm}
\end{figure*}

Table~\ref{tab:main_results} presents the main results on OLMo-2 models across both benchmarks.

\noindent\textbf{QuCo-RAG Achieves Significant Improvements over Baselines.}
Across all model scales and datasets, QuCo-RAG consistently outperforms the strongest baselines by significant margins. On OLMo-2-7B, QuCo-RAG achieves 32.7 EM on 2WikiMultihopQA and 35.3 EM on HotpotQA, surpassing the best baseline by +7.4 and +5.6 points respectively. The improvements become even more pronounced with larger models: OLMo-2-13B shows gains of +12.0 EM on 2WikiMultihopQA, while OLMo-2-32B achieves +10.8 EM improvements on HotpotQA. These results demonstrate that grounding retrieval decisions in corpus statistics provides a fundamentally more reliable signal than model-internal uncertainty measures.

\noindent\textbf{Internal-Signal Methods Show Inconsistent Performance.}
Methods relying on model-internal signals (FLARE, DRAGIN, ETC, SeaKR) show highly variable results across settings. 
For instance, ETC achieves second-best performance in some configurations, yet underperforms even simple SR-RAG in others. 
DRAGIN achieves only 17.5--19.5 EM on HotpotQA across all model sizes, substantially underperforming SR-RAG.
This inconsistency stems from the fundamental unreliability of internal uncertainty signals. A detailed case study is provided in Appendix~\ref{sec:case_study}.

% (e.g., 4.6 EM lower than SR-RAG on HotpotQA with OLMo-2-7B)

% Required packages (add to preamble):
% \usepackage{booktabs}
% \usepackage[table]{xcolor}
% \usepackage{adjustbox}
% \definecolor{lightblue}{RGB}{173,216,230}
% \definecolor{darkred}{RGB}{178,34,34}

\begin{table}[t]
    \centering
    \caption{Transferability to other model families (EM scores). HPQA: HotpotQA. `-' indicates the method is not applicable due to API limitations. Full results with F1 score are in Appendix~\ref{sec:appendix_transferability}.}

    % \caption{Transferability to other model families (EM scores). Full results with F1 in Appendix.}
    \label{tab:other_models}
    \begin{adjustbox}{max width=0.42\textwidth}
    \begin{tabular}{l cc cc}
        \toprule
        & \multicolumn{2}{c}{\textbf{Qwen2.5-32B}} & \multicolumn{2}{c}{\textbf{Llama-3-8B}} \\
        \cmidrule(lr){2-3} \cmidrule(lr){4-5}
        Method & 2Wiki & HPQA & 2Wiki & HPQA \\
        \midrule
        Wo-RAG & 26.4 & 21.6 & 29.5 & 20.3 \\
        SR-RAG & 23.0 & 31.0 & 12.9 & 22.7 \\
        FS-RAG & \underline{35.9} & \underline{38.6} & 28.8 & 27.0 \\
        FLARE & 26.4 & 24.1 & 26.6 & 22.2 \\
        DRAGIN & 28.8 & 22.2 & 27.9 & 20.0 \\
        ETC & 31.5 & 21.7 & 29.9 & 24.1 \\
        SeaKR & 22.4 & 26.7 & \underline{33.5} & \underline{33.5} \\
        \midrule
        \rowcolor{lightblue!20}
        QuCo-RAG & \textbf{50.0} & \textbf{41.6} & \textbf{38.4} & \textbf{36.2} \\
        \rowcolor{lightblue!20}
        \textit{Improv.} & \textcolor{darkred}{\textbf{+14.1}} & \textcolor{darkred}{\textbf{+3.0}} & \textcolor{darkred}{\textbf{+4.9}} & \textcolor{darkred}{\textbf{+2.7}} \\
        \midrule
        \addlinespace[4pt]
        \midrule
        & \multicolumn{2}{c}{\textbf{GPT-4.1}} & \multicolumn{2}{c}{\textbf{GPT-5-chat}} \\
        \cmidrule(lr){2-3} \cmidrule(lr){4-5}
        Method & 2Wiki & HPQA & 2Wiki & HPQA \\
        \midrule
        Wo-RAG & 54.7 & \underline{40.1} & 50.1 & 37.7 \\
        SR-RAG & \underline{60.0} & 38.8 & \underline{51.0} & \underline{42.9} \\
        FS-RAG & 59.5 & 25.9 & 47.3 & 19.0 \\
        FLARE & 49.8 & 38.7 & - & - \\
        Web-Tool & 42.9 & 8.9 & 48.3 & 19.8 \\
        \midrule
        \rowcolor{lightblue!20}
        QuCo-RAG & \textbf{64.6} & \textbf{48.2} & \textbf{59.7} & \textbf{48.4} \\
        \rowcolor{lightblue!20}
        \textit{Improv.} & \textcolor{darkred}{\textbf{+4.6}} & \textcolor{darkred}{\textbf{+8.1}} & \textcolor{darkred}{\textbf{+8.7}} & \textcolor{darkred}{\textbf{+5.5}} \\
        \bottomrule
    \end{tabular}
    \end{adjustbox}
    \vspace{-0.24cm}
\end{table}

\subsection{Transferability to Other Models (RQ2)}
\label{sec:transferability}

A critical question for corpus-based methods is whether they generalize to models whose training data is proprietary or undisclosed. 
We evaluate QuCo-RAG on Qwen2.5, Llama-3, and GPT model families, using the OLMo-2 corpus as a \textit{proxy corpus} for their knowledge distributions (Table~\ref{tab:other_models}).

\noindent\textbf{Effectiveness Across Model Families.}
QuCo-RAG demonstrates remarkable transferability, consistently outperforming all baselines across model families.
On open-weight models, it achieves substantial gains; notably, for Qwen2.5-32B on 2WikiMultihopQA, our method obtains a +14.1 EM improvement over the strongest baseline.
This trend extends to proprietary models: QuCo-RAG improves GPT-5-chat by +8.7 EM on 2WikiMultihopQA and +5.5 EM on HotpotQA.
Conversely, GPT models with the web search tool perform substantially worse than even the no-retrieval baseline, likely because web search returns noisy results and GPT's search capability is not optimized for complex retrieval demands.
% likely due to noisy web results not optimized for complex retrieval demands.

\noindent\textbf{Why Proxy Corpus Works.}
The effectiveness of cross-model transfer validates our hypothesis that web-scale pre-training corpora share substantial overlap~\cite{soldaini2024dolma,li2024datacomp}. Factual knowledge is largely drawn from common sources such as Common Crawl, Wikipedia, and books, making frequency and co-occurrence statistics from one comprehensive corpus a reliable proxy for others. This property renders QuCo-RAG practically \textit{model-agnostic}.

\subsection{Efficiency Analysis (RQ3)}
\label{sec:efficiency}

\begin{figure}[t]
    \centering
    \includegraphics[width=0.98\linewidth]{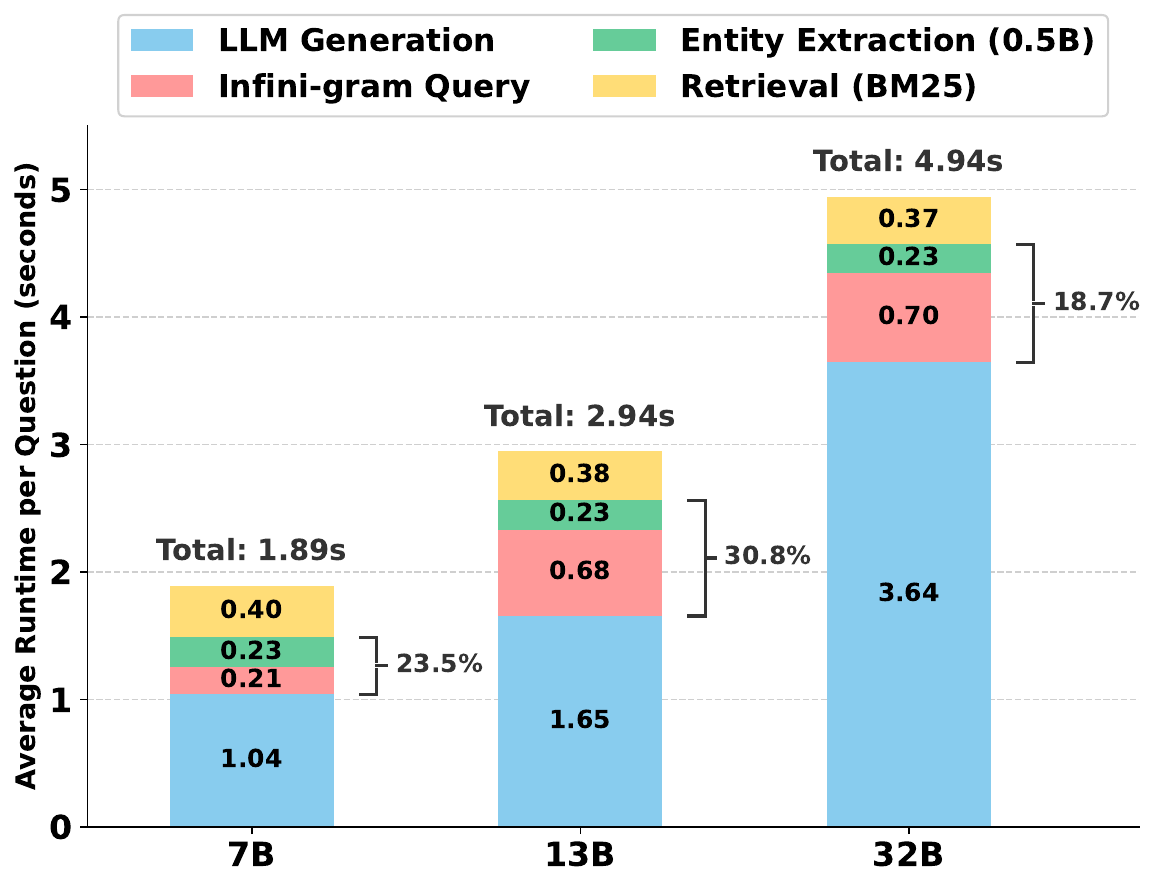}
    \caption{Average runtime breakdown per question for QuCo-RAG components across OLMo-2 model sizes on 2WikiMultihopQA.}
    \label{fig:runtime_breakdown}
    \vspace{-4mm}
\end{figure}

Figure~\ref{fig:efficiency} illustrates the efficiency-performance trade-off of different methods. QuCo-RAG achieves the highest EM (35.0) while consuming only 87 tokens and 1.84 LLM calls on average, both the lowest among dynamic RAG methods. FS-RAG and DRAGIN consume 2--4$\times$ more tokens yet achieve substantially lower performance, while SeaKR incurs excessive LLM calls (10.28) due to repeated hidden-state uncertainty estimation. As shown in Figure~\ref{fig:efficiency}(c), QuCo-RAG triggers only 1.70 retrievals per question on average, demonstrating precise corpus-grounded detection. 
Notably, no baseline falls in the green region (higher EM with fewer retrievals than QuCo-RAG), while methods like FLARE and FS-RAG fall in the red region, performing worse than Wo-RAG despite frequent retrieval.
Regarding runtime, Figure~\ref{fig:runtime_breakdown} shows that LLM generation dominates (55--74\% of total time), while corpus-based detection introduces modest overhead, demonstrating favorable scaling for deployment.

% (Infini-gram queries: 11--23\%; entity extraction: 0.23s) (55--74\%)

\section{Analysis and Discussion}

We provide additional analyses including ablation studies, generalization studies, and performance breakdown by entity frequency. Threshold sensitivity analysis is provided in Appendix~\ref{sec:threshold_sensitivity}, and an analysis of Stage~2's co-occurrence verification, 
including a false negative breakdown and an optional relation-aware extension, is provided in 
Appendix~\ref{sec:cooc_analysis}.

\subsection{Ablation Studies}

Table~\ref{tab:ablation} examines the contribution of each detection stage. Removing Pre-Generation Knowledge Assessment (w/o Initial Check) reduces EM by 2.5 points, confirming that identifying rare entities in the question is valuable for the initial response. Removing Runtime Claim Verification (w/o Runtime Check) causes a larger drop of 5.1 EM points, demonstrating that co-occurrence verification is the more critical component.
Interestingly, even w/o Runtime Check (Initial Check only) outperforms SR-RAG by 3.9 EM while triggering fewer retrievals (0.76 vs. 1.00). This suggests that selective retrieval based on entity frequency can be more effective than always-retrieve strategies at the pre-generation stage---not all questions benefit equally from retrieval, and frequency-based detection provides a useful signal for prioritizing retrieval.

\begin{table}[t]
    \centering
    \caption{Ablation study on two-stage detection (2WikiMultihopQA, OLMo-2-7B). \#Ret.: average retrieval count per question.}
    \label{tab:ablation}
    \begin{adjustbox}{max width=0.93\columnwidth}
    \begin{tabular}{lccc}
        \toprule
        % \rowcolor{cyan!10}
        \textbf{Configuration} & \textbf{EM} & \textbf{F1} & \textbf{\#Ret.} \\
        \midrule
        QuCo-RAG (Full) & 32.7 & 41.1 & 2.61 \\
        \quad w/o Initial Check & 30.2\textsubscript{\textcolor{darkred}{\textbf{-2.5}}} & 38.0\textsubscript{\textcolor{darkred}{\textbf{-3.1}}} & 1.82 \\
        \quad w/o Runtime Check & 27.6\textsubscript{\textcolor{darkred}{\textbf{-5.1}}} & 35.6\textsubscript{\textcolor{darkred}{\textbf{-5.5}}} & 0.76 \\
        \midrule
        \multicolumn{4}{l}{\textit{Baselines}} \\
        \quad SR-RAG & 23.7 & 30.7 & 1.00 \\
        \quad Wo-RAG & 20.1 & 26.4 & 0.00 \\
        \bottomrule
    \end{tabular}
    \end{adjustbox}
     \vspace{-0.4cm}
\end{table}

\subsection{Generalization Studies}

To evaluate whether QuCo-RAG generalizes beyond short-answer open-domain QA, we test on two additional dimensions: long-form generation and domain-specific biomedical QA.

\paragraph{Long-Form Generation.}
We evaluate on ASQA~\citep{stelmakh-etal-2022-asqa}, a long-form QA benchmark where models must generate comprehensive multi-sentence answers (averaging 8--10 sentences) to ambiguous questions with multiple valid interpretations.
Following the original benchmark setup, we report ROUGE-L and the composite disambiguation-ROUGE (DR) metric:
\begin{equation}
\text{DR} = \sqrt{\text{Disambig-F1} \times \text{ROUGE-L}}.
\end{equation}
We additionally report LLM\_DR, which replaces the lexical Disambig-F1 in DR with a GPT-4o-based semantic correctness score~\citep{li2025generation}, to better capture answer quality beyond surface-level overlap.
Results are shown in Table~\ref{tab:asqa}.

\begin{table}[h]
\centering
\small
\begin{tabular}{lcccc}
\toprule
\textbf{Method} & \textbf{ROUGE-L} & \textbf{DR} & \textbf{LLM\_DR} & \textbf{\#Ret.} \\
\midrule
Wo-RAG   & 21.9 & 22.7 & 16.9 & 0.00 \\
SR-RAG   & 22.7 & \underline{26.5} & \underline{21.8} & 1.00 \\
FS-RAG   & 20.3 & 23.5 & 20.9 & 11.15 \\
FLARE    & 19.1 & 22.5 & 18.7 & 4.53 \\
DRAGIN   & 16.9 & 19.6 & 15.3 & 3.62 \\
ETC      & \underline{22.8} & 25.1 & 19.9 & 2.92 \\
\midrule
\rowcolor{lightblue!20}
QuCo-RAG & \textbf{28.9} & \textbf{28.5} & \textbf{23.3} & 1.72 \\
\rowcolor{lightblue!20}
\textit{Improv.} & \textcolor{darkred}{\textbf{+6.1}} & \textcolor{darkred}{\textbf{+2.0}} & \textcolor{darkred}{\textbf{+1.5}} & --- \\
\bottomrule
\end{tabular}
\caption{Long-form generation results on ASQA (OLMo-2-7B). \#Ret.: average retrieval count per question.}
\label{tab:asqa}
\end{table}

QuCo-RAG achieves the highest scores on both standard metrics (ROUGE-L, DR) and the LLM-based evaluation (LLM\_DR) with only 1.72 retrievals per question.
Notably, aggressive retrieval strategies (FS-RAG with 11.15, FLARE with 4.53 retrievals) actually degrade ROUGE-L below Wo-RAG, suggesting that excessive retrieval may introduce noise that harms long-form coherence~\citep{Cuconasu_2024}.
In contrast, QuCo-RAG's selective, corpus-grounded retrieval avoids this pitfall, demonstrating that our paradigm extends effectively from short-answer to long-form generation settings.

\paragraph{Domain Generalization.}
We further test on PubMedQA~\citep{jin2019pubmedqa}, a biomedical QA benchmark where models answer research questions based on biomedical literature.
Following~\citet{xiong-etal-2024-benchmarking}, we use PubMed abstracts and medical textbooks~\cite{jin2020disease} as the retrieval corpus $\mathcal{C}$ and report accuracy following the standard benchmark setup~\cite{wu-etal-2025-medical}.
Notably, we retain the same OLMo-2 pre-training corpus as the statistical signal source $\mathcal{P}$, without any domain-specific adaptation.
As shown in Table~\ref{tab:pubmedqa}, QuCo-RAG achieves the best accuracy (66.4\%) while maintaining high efficiency (0.93 retrievals, 54.9 tokens per question). Internal-signal methods show limitations in this specialized domain. FLARE achieves decent accuracy (63.4\%) but at 9$\times$ the token cost of ours (516.8 vs. 54.9 tokens), as its probability-based triggering becomes overly sensitive to domain-specific biomedical terminology. Conversely, DRAGIN and ETC perform no better than Wo-RAG, likely because their entropy-based uncertainty formulations fail to generalize across domains. QuCo-RAG avoids these pitfalls: large-scale pre-training corpora provide broad coverage of biomedical knowledge, and zero co-occurrence reliably signals high uncertainty regardless of domain.

\begin{table}[h]
    \centering
    % \vspace{-0.4cm}
    \caption{Domain generalization on PubMedQA (OLMo-2-7B). $\Delta$Acc: improvement over Wo-RAG; \#Tok.: average token consumption per question.}
    \label{tab:pubmedqa}
    \begin{adjustbox}{max width=0.83\columnwidth}
    \begin{tabular}{lcccc}
        \toprule
        Method & Acc & $\Delta$Acc & \#Ret. & \#Tok. \\
        \midrule
        Wo-RAG & 55.2 & 0.0 & 0.00 & 40.3 \\
        FS-RAG & 61.1 & +5.9 & 5.74 & 436.1 \\
        FLARE & \underline{63.4} & \underline{+8.2} & 2.79 & 516.8 \\
        DRAGIN & 55.2 & 0.0 & 1.69 & 139.0 \\
        ETC & 55.0 & -0.2 & 0.25 & 58.8 \\
        \midrule
        \rowcolor{lightblue!20}
        \textbf{QuCo-RAG} & \textbf{66.4} & \textbf{+11.2} & 0.93 & 54.9 \\
        \bottomrule
    \end{tabular}
    \end{adjustbox}
     \vspace{-0.4cm}
\end{table}

\subsection{Performance Across Entity Frequency}

\begin{figure}[t]
    \centering
    \includegraphics[width=\columnwidth]{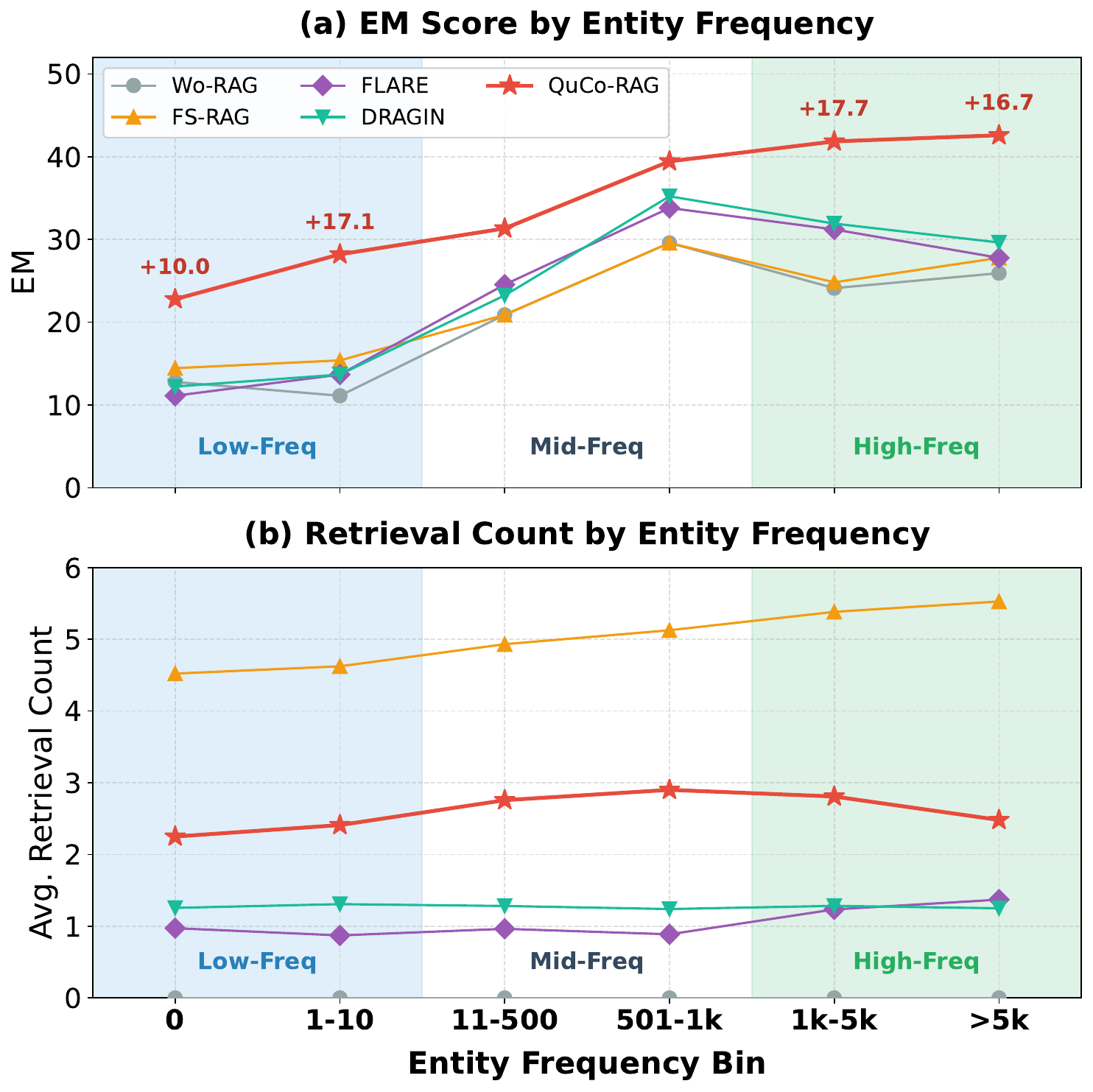}
    \caption{Performance stratified by entity frequency bins on 2WikiMultihopQA (OLMo-2-7B).}
    \label{fig:frequency_analysis}
    \vspace{-0.5cm}
\end{figure}

To understand how different methods handle knowledge of varying prevalence, we group questions by how often their entities appear in the pre-training corpus. Figure~\ref{fig:frequency_analysis} shows EM scores and retrieval counts across frequency bins. Full numerical results are provided in Appendix Table~\ref{tab:frequency_breakdown}. Overall, all methods perform worse in low-frequency bins, confirming that entity frequency correlates with model reliability.
% and rare entities pose greater hallucination risks
In \textbf{low-frequency bins (0--10)}, QuCo-RAG demonstrates dominant performance, outperforming Wo-RAG by 10--17 EM points, while DRAGIN and FLARE achieve nearly identical performance to Wo-RAG despite triggering retrievals, suggesting that models lack sufficient signal to recognize uncertainty on rare entities.
In \textbf{mid-frequency bins (11--1k)}, the gap narrows as internal-signal methods become competitive, likely because mid-frequency entities place models in a ``partially learned'' state where model-internal signals become relatively effective.
In \textbf{high-frequency bins (>1k)}, an interesting divergence emerges: baselines exhibit performance degradation while QuCo-RAG continues to improve. For internal-signal methods, the decline is likely due to overconfidence, failing to trigger retrieval even when generating wrong claims. In contrast, QuCo-RAG benefits from richer knowledge coverage: high-frequency entities have more documented co-occurrences in the corpus, making co-occurrence statistics more reliable.

% Meanwhile, QuCo-RAG's retrieval count decreases in this region, as models are more likely to generate correct claims for common entities.

\subsection{Broader Impact and Future Directions}

Our work establishes corpus statistics as an objective alternative to model-internal uncertainty signals; while this paper focuses on retrieval triggering in RAG systems, the paradigm shift opens several promising avenues in AI safety and robustness.

\noindent\textbf{Enabling Trustworthy AI Applications.}
Our experiments establish that corpus statistics offer a reliable uncertainty measure. This reliability is critical not only for RAG but also for broader safety-critical tasks, such as \textit{selective answering}, where models can decline to answer when evidential support is absent, and \textit{correctness prediction}, where corpus statistics provide well-grounded confidence scores for generated claims.

\noindent\textbf{From Inference-Time Intervention to Data-Centric AI.} 
Our corpus statistics analysis precisely identifies the model's knowledge gaps. This signal can inform \textit{training data curation}: rather than only compensating for gaps at inference time via retrieval, developers can proactively collect data for low-frequency entities during continued pre-training or post-training. Similarly, corpus statistics can guide \textit{synthetic data filtering}, where LLM-generated training examples are verified against corpus statistics before inclusion, and \textit{model editing} by distinguishing facts that require targeted injection from those already reliably learned.

\noindent\textbf{Extensions of the Paradigm.}
Several directions merit exploration: (1) multilingual verification through cross-lingual statistics; (2) temporal dynamics via time-stamped corpora for evolving knowledge; (3) extension beyond entities to events, relations, and numerical claims~\cite{min-etal-2024-exploring,10.1007/978-3-031-47240-4_19, wei2026epiqal,10.1145/3746027.3755764}; and (4) integration into agentic systems~\citep{huang2026rethinkingmemorymechanismsfoundation,wu2026gamhierarchicalgraphbasedagentic, ho2025arcmemo} as a self-verification tool that agents invoke before acting on generation.

\noindent\textbf{Theoretical Foundations.}
Our transferability results raise fundamental questions: 
(1) can we formally characterize the relationship between corpus statistics and model knowledge?
(2) can we formalize information-theoretic bounds on hallucination probability given corpus statistics? These questions connect to broader debates on memorization versus generalization in LLMs.

\section{Conclusion}
\label{sec:Conclusion}
We propose QuCo-RAG, a dynamic RAG framework that quantifies uncertainty from pre-training corpus statistics rather than poorly calibrated model-internal signals. QuCo-RAG achieves state-of-the-art performance on multi-hop QA benchmarks while maintaining superior efficiency, transfers effectively to models with undisclosed training data (Llama-3, Qwen2.5, GPT-4.1/5-chat), and generalizes robustly to long-form generation and biomedical QA. These results establish corpus-grounded verification as a principled, practically model-agnostic paradigm for dynamic RAG.

\section*{Limitations}

\noindent\textbf{(1) Lexical Matching Constraints.}
Our co-occurrence verification relies on exact lexical matching of entity surface forms. This may lead to false positive retrieval triggers when two genuinely related entities co-occur in the corpus under alternative names or aliases (e.g., ``NYC'' vs. ``New York City''), yet show zero co-occurrence for the specific surface forms extracted from the generated text. However, we argue this limitation is acceptable in practice due to the \textit{asymmetric risk} inherent in RAG systems: the cost of an unnecessary retrieval (slightly increased latency) is far lower than that of an undetected hallucination (incorrect output). Our conservative strategy, triggering retrieval when in doubt, thus errs on the side of caution. 
Moreover, given the massive scale of the pre-training corpus, genuinely related entities typically co-occur in some form, mitigating alias-induced false alarms. Future work could incorporate entity linking or canonicalization techniques to further reduce unnecessary retrievals while maintaining verification recall.

\noindent\textbf{(2) Temporal Limitations of Static Corpora.}
Our approach inherits the temporal limitations of static pre-training corpora. A corpus indexed at a particular point in time cannot provide meaningful statistics for entities or events that emerge afterward (e.g., a 2024 corpus cannot verify claims about 2025 sports results or newly founded organizations). This limitation can be addressed through periodic corpus updates and index maintenance, which modern infrastructure like Infini-gram supports efficiently.

\section*{Acknowledgments}
This work is supported by the National Science Foundation (NSF) Grant \#2312862, NSF-Simons SkAI Institute, NSF CAREER \#2440542, NSF \#2533996, National Institutes of Health (NIH) \#R01AG091762, NSF ACCESS Computing Resources, NAIRR, a Google Research Scholar Award, and Cisco gift grant.

This research used the Delta advanced computing and data resource at the National Center for Supercomputing Applications (NCSA) through allocations CIS260189 and CIS250364 from the Advanced Cyberinfrastructure Coordination Ecosystem: Services \& Support (ACCESS) program~\citep{10.1145/3569951.3597559}. Delta is supported by the National Science Foundation (award OAC 2005572) and the State of Illinois, and is a joint effort of the University of Illinois Urbana-Champaign and its NCSA.
We also thank the National Research Platform (NRP)~\citep{10.1145/3708035.3736060}, 
supported by the U.S.\ National Science Foundation, for providing 
computational resources.

% % \section*{Acknowledgments}

% % \section*{Ethical Considerations}

\bibliography{custom}

\appendix

\newpage
\section{Appendix}

\subsection{Additional Implementation Details}
\label{app:prompts}

\paragraph{Generation Settings and Prompts.}
In our experiments, all open-source models use greedy decoding with a 128-token generation limit per step, and GPT models use default parameters via API calls.
For generation, we employ 6-to-8-shot Chain-of-Thought prompting~\cite{wei2022chain}, adopting templates from \citet{trivedi2023interleaving} and \citet{jiang-etal-2023-active}. We use 6 few-shot examples for 2WikiMultihopQA and 8 for HotpotQA, consistent with prior work. 
The full prompt template is provided in Table~\ref{tab:prompt_template}. We use the Wikipedia dump from~\citet{karpukhin-etal-2020-dense} as our external corpus $\mathcal{C}$, which contains approximately 21 million passages.

\begin{table}[h!]
\small
\begin{tcolorbox}
\textcolor{blue}{\textbf{Few-shot Examples:}}
\\\\
\textbf{Question:} When did the director of film Hypocrite (Film) die?
\\
\textbf{Answer:} The film Hypocrite was directed by Miguel Morayta. Miguel Morayta died on 19 June 2013. So the answer is 19 June 2013.
\\\\
\textit{[... 5--7 more demonstrations ...]}
\\\\
\textcolor{blue}{\textbf{Retrieved Context (if available):}}
\\\\
Background information that may be potentially useful in addressing your question:
\\
{[}1{]} \textit{<retrieved document 1>}
\\
{[}2{]} \textit{<retrieved document 2>}
\\
{[}3{]} \textit{<retrieved document 3>}
\\\\
\textcolor{blue}{\textbf{Instruction:}}
\\\\
Please answer the following questions. The format of the answers should be the same as the examples given before. Specifically, you need to think through the answer to this question step by step. Each sentence should only present a fact statement. Avoid using pronouns like He/She/It or possessive pronouns like His/Her/Its, but instead use specific names. At the end of your answer, use ``So the answer is'' to provide your answer.
\\\\
\textcolor{blue}{\textbf{Question:}} \textit{<input question>}
\end{tcolorbox}
\caption{Prompt template used for multi-hop QA experiments. Retrieved context is prepended when retrieval is triggered.}
\label{tab:prompt_template}
\end{table}

\paragraph{Answer Extraction.}
Following standard practice adopted by all baselines~\cite{su-etal-2024-dragin,jiang-etal-2023-active,li2025modeling}, we extract the final answer via regex matching the pattern ``So the answer is''. Extraction success rates are 91.5--99.0\% on 2WikiMultihopQA and 97.9--99.3\% 
on HotpotQA across OLMo-2 model sizes (7B/13B/32B), with larger models exhibiting higher compliance. For the remaining unmatched cases, we append ``So the answer is'' and prompt the model to continue generation 
to elicit the final answer. This procedure is identical across all methods.

\paragraph{Local Deployment Resources.}
In our experiments, we query the publicly hosted Infini-gram API.
For local deployment, the Infini-gram suffix array index for the OLMo-2 pre-training corpus (${\sim}4$ trillion tokens) requires approximately 28~TB of disk storage and ${\sim}300$~MB query-time RAM, with no GPU needed.
The index construction is a one-time cost (${\sim}100$~h on 128 CPUs); once built, it serves all subsequent queries across all models.
For reduced storage, Infini-gram Mini~\citep{xu-etal-2025-infini} compresses the index to $0.44\times$ the corpus size (${\sim}4$~TB) via FM-index.

\subsection{Case Study: Uncertainty Quantification in Action}
\label{sec:case_study}

Table~\ref{tab:case_study} presents a detailed case study demonstrating how QuCo-RAG quantifies uncertainty through corpus statistics to detect and correct hallucinations that baseline methods miss. 
In this multi-hop question, all baselines fail for distinct reasons: Wo-RAG hallucinates without any correction mechanism; SR-RAG retrieves correct director information but cannot perform follow-up retrieval for the mother; FLARE detects some uncertainty but its query contains the hallucinated director name ``Igor Maslennikov,'' leading to retrieval of irrelevant documents; DRAGIN's internal signals mark this completely fabricated director as low uncertainty, exemplifying the confident hallucination problem, and its subsequent query still contains the error, reinforcing the mistake.
In contrast, QuCo-RAG succeeds through the coordination of two stages: Stage 1 identifies ``Polish-Russian War'' as a low-frequency entity, triggering initial retrieval that grounds the model to generate the correct director ``Xawery Żuławski.'' Stage 2 then catches the hallucinated mother ``Anna Żuławski'' via zero co-occurrence (the two entities never appear together in the corpus), triggering targeted retrieval with a hallucination-free query ``Xawery Żuławski mother'' that yields the correct answer.

\begin{table*}[t]
\small
\centering
\caption{Case study comparison. \textcolor{darkred}{\textbf{Red}} indicates hallucinated/incorrect content; \textcolor{green!50!black}{\textbf{green}} indicates correct content. Only QuCo-RAG produces the correct answer through corpus-grounded uncertainty quantification.}
\label{tab:case_study}
\begin{adjustbox}{max width=\textwidth}
\begin{tabular}{l|p{3.4cm}|p{3.0cm}|p{2.4cm}|p{2.2cm}|p{3.4cm}}
\toprule
\multicolumn{6}{l}{\textbf{Question:} \textit{Who is the mother of the director of film Polish-Russian War?}} \\
\multicolumn{6}{l}{\textbf{Ground Truth:} \textbf{Małgorzata Braunek} \quad (Polish-Russian War (film) $\rightarrow$ Director: Xawery Żuławski $\rightarrow$ Mother: Małgorzata Braunek)} \\
\midrule
\textbf{Method} & \textbf{Initial Generation} & \textbf{Uncertainty Signal} & \textbf{Retrieval Query} & \textbf{Final Answer} & \textbf{Analysis} \\
\midrule
Wo-RAG & 
``...directed by \textcolor{darkred}{\textbf{Igor Maslennikov}}. His mother is \textcolor{darkred}{\textbf{Natalia Maslennikova}}.'' &
N/A &
N/A &
\textcolor{darkred}{\textbf{Natalia Maslennikova}} &
No retrieval mechanism to correct hallucinated director. \\
\midrule
SR-RAG &
``...directed by \textcolor{green!50!black}{\textbf{Xawery Żuławski}}. No information about his mother.'' &
N/A (retrieves once before generation) &
Original question &
\textcolor{darkred}{\textbf{unknown}} &
Single-round retrieval insufficient for multi-hop reasoning. \\
\midrule
FLARE &
``...directed by \textcolor{darkred}{\textbf{Igor Maslennikov}}. His mother is \textcolor{darkred}{\textbf{Svetlana}}.'' &
Triggered at sentence-level (probability below threshold) &
\textcolor{darkred}{\textbf{``Igor Maslennikov...''}} &
\textcolor{darkred}{\textbf{unknown}} &
Query included hallucinated director; retrieved irrelevant documents. \\
\midrule
DRAGIN &
``...directed by \textcolor{darkred}{\textbf{Igor Maslennikov}}. His mother is \textcolor{darkred}{\textbf{Natalia Maslennikova}}.'' &
Triggered at token ``Natalia'' (entropy-based); \newline \textcolor{darkred}{\textbf{wrong director}} marked as low uncertainty &
\textcolor{darkred}{\textbf{``Igor Maslennikov mother''}} &
\textcolor{darkred}{\textbf{Natalia Maslennikova}} &
Confident hallucination: internal signals failed to flag the wrong director; query contained error, reinforcing mistake. \\
\midrule
\rowcolor{lightblue!20}
QuCo-RAG &
S1: ``...directed by \textcolor{green!50!black}{\textbf{Xawery Żuławski}}.'' \newline
S2: ``...mother is \textcolor{darkred}{\textbf{Anna Żuławski}}.'' &
\textbf{Stage 1:} Low entity freq. $\rightarrow$ retrieval \newline
\textbf{Stage 2:} Co-occurrence = \textbf{0} $\rightarrow$ high uncertainty &
\textbf{Stage 1:} Original question \newline
\textbf{Stage 2:} \textcolor{green!50!black}{\textbf{``Xawery Żuławski mother''}}  &
\textcolor{green!50!black}{\textbf{Małgorzata Braunek}} &
Stage 1 ensured correct director via initial retrieval; Stage 2 caught hallucinated mother via zero co-occurrence. \\
\bottomrule
\end{tabular}
\end{adjustbox}
\end{table*}

\subsection{Triplet Extractor Training Examples}
\label{app:triplet_examples}

Table~\ref{tab:triplet_examples} shows representative examples from our triplet extractor training data. Each example consists of an input sentence and the extracted output. If the input sentence contains meaningful factual knowledge, the output consists of knowledge triplets in the format (head entity, relation, tail entity); otherwise, the output is empty.
We prioritize extracting triplets where the tail entity is a named entity (person, location, organization, date) rather than generic descriptors, as these are more amenable to corpus co-occurrence verification. Non-factual statements such as reasoning conclusions (e.g., sentences starting with "Thus" or "Therefore") return empty outputs since they do not introduce new verifiable facts.

\begin{table*}[ht]
\small
\centering
\caption{Examples of triplet extractor training data. The model extracts factual triplets from declarative sentences, partial triplets from questions (since the answer is unknown), and returns empty for non-factual statements.}
\label{tab:triplet_examples}
\begin{tabular}{p{5.8cm}|p{5.8cm}}
\toprule
\textbf{Input Sentence} & \textbf{Extracted Output} \\
\midrule
\rowcolor{gray!15}
\multicolumn{2}{l}{\textbf{\textit{Declarative sentences with factual knowledge:}}} \\
\midrule
Kumbasaram was released in 2017. & 
[["Kumbasaram", "released in", "2017"]] \\
\midrule
Beowulf \& Grendel was directed by Sturla Gunnarsson. & 
[["Beowulf \& Grendel", "directed by", "Sturla Gunnarsson"]] \\
\midrule
Coulson Wallop's father, Nigel Wallop, studied at Eton College. & 
[["Coulson Wallop", "father", "Nigel Wallop"], ["Nigel Wallop", "studied at", "Eton College"]] \\
\midrule
\rowcolor{gray!15}
\multicolumn{2}{l}{\textbf{\textit{Questions (answer unknown, extract partial triplets):}}} \\
\midrule
Which film came out first, Kumbasaram or Mystery Of The 13th Guest? & 
[["Kumbasaram", "came out"], ["Mystery of the 13th Guest", "came out"]] \\
\midrule
Where did Diane Meyer Simon's husband graduate from? & 
[["Diane Meyer Simon", "husband"]] \\
\midrule
\rowcolor{gray!15}
\multicolumn{2}{l}{\textbf{\textit{Non-factual statements (reasoning conclusions):}}} \\
\midrule
Thus, Kumbasaram came out first. & 
[] \\
\midrule
Therefore, Robert Enrico, the director of The Woman Thou Gavest Me, was born first. & 
[] \\
\bottomrule
\end{tabular}
\end{table*}

\subsection{Triplet Extractor Evaluation}
\label{app:extractor_eval}
 
We evaluate the distilled 0.5B extractor on 1,000 randomly sampled held-out instances (disjoint from the 40K training set).
We extract unique head/tail entities from predicted and ground-truth triplets and compute exact matching.
On 739 factual sentences, the extractor achieves 89.9\% entity-level F1 (Precision 93.3\%, Recall 86.8\%); on 261 non-factual sentences, it correctly predicts empty output 81.8\% of the time.

More importantly, we conduct an end-to-end ablation by replacing the 0.5B extractor with the GPT-4o-mini teacher in the full QuCo-RAG pipeline (Table~\ref{tab:extractor_ablation}).
The 0.5B model achieves comparable or slightly higher EM than GPT-4o-mini across both benchmarks, confirming that extraction quality is not a bottleneck.
The slight advantage of the 0.5B model is likely because the distilled model, after full-parameter fine-tuning, produces more consistent output formatting, whereas GPT-4o-mini occasionally generates irregular triplet structures.
 
\begin{table}[h]
\centering
\small
\begin{tabular}{l cccc}
\toprule
& \multicolumn{2}{c}{\textbf{2Wiki}} & \multicolumn{2}{c}{\textbf{HPQA}} \\
\cmidrule(lr){2-3} \cmidrule(lr){4-5}
Triplet Extractor & EM & F1 & EM & F1 \\
\midrule
GPT-4o-mini (teacher)       & 40.6 & 48.1 & 34.0 & 44.8 \\
QuCo-extractor-0.5B (ours)  & \textbf{41.7} & \textbf{49.1} & \textbf{35.0} & \textbf{46.8} \\
\bottomrule
\end{tabular}
\caption{End-to-end ablation of extractor choice (OLMo-2-13B). 2Wiki: 2WikiMultihopQA; HPQA: HotpotQA.}
\label{tab:extractor_ablation}
\end{table}

\subsection{Full Results for Transferability Experiments}
\label{sec:appendix_transferability}

Table~\ref{tab:other_models_comparison_full} presents the complete results (EM and F1) for the transferability experiments discussed in Section~\ref{sec:transferability}. The main paper reports only EM scores for brevity. Across all model families (Qwen2.5-32B, Llama-3-8B, GPT-4.1, and GPT-5-chat), QuCo-RAG consistently achieves the best performance on both metrics. The F1 improvements follow similar patterns to EM, confirming that QuCo-RAG's gains are robust.

\begin{table}[h!]
    \centering
    \caption{Comparison of different RAG methods on 2WikiMultihopQA and HotpotQA benchmarks.}
    \label{tab:other_models_comparison_full}
    \adjustbox{max width=0.4\textwidth}{
    \begin{tabular}{lcccc}
        \toprule
        & \multicolumn{2}{c}{2Wiki} & \multicolumn{2}{c}{HotpotQA} \\
        \cmidrule(lr){2-3} \cmidrule(lr){4-5}
        Method & EM & F1 & EM & F1 \\
        \midrule
        \multicolumn{5}{l}{\textbf{Qwen2.5-32B-Instruct}} \\
        \midrule
        Wo-RAG & 26.4 & 33.6 & 21.6 & 32.4 \\
        SR-RAG & 23.0 & 31.8 & 31.0 & 41.7 \\
        FS-RAG & \underline{35.9} & \underline{45.3} & \underline{38.6} & \underline{49.6} \\
        FLARE  & 26.4 & 33.3 & 24.1 & 33.5 \\
        DRAGIN & 28.8 & 36.9 & 22.2 & 32.4 \\
        ETC    & 31.5 & 40.2 & 21.7 & 32.0 \\
        SeaKR  & 22.4 & 31.3 & 26.7 & 37.5 \\
        \rowcolor{lightblue!20} QuCo-RAG & \textbf{50.0} & \textbf{58.9} & \textbf{41.6} & \textbf{55.1} \\
        
        \midrule
        \multicolumn{5}{l}{\textbf{Llama-3-8B-Instruct}} \\
        \midrule
        Wo-RAG & 29.5 & 37.7 & 20.3 & 31.4 \\
        SR-RAG & 12.9 & 29.2 & 22.7 & 35.4 \\
        FS-RAG & 28.8 & 36.8 & 27.0 & 38.5 \\
        FLARE  & 26.6 & 35.1 & 22.2 & 31.5 \\
        DRAGIN & 27.9 & 36.7 & 20.0 & 31.9 \\
        ETC    & 29.9 & 39.2 & 24.1 & 35.1 \\
        SeaKR  & \underline{33.5} & \underline{40.4} & \underline{33.5} & \underline{46.0} \\
        \rowcolor{lightblue!20} QuCo-RAG & \textbf{38.4} & \textbf{46.6} & \textbf{36.2} & \textbf{48.7} \\
        
        \midrule
        \multicolumn{5}{l}{\textbf{GPT-4.1}} \\
        \midrule
        Wo-RAG & 54.7 & 69.9 & \underline{40.1} & \underline{56.1} \\
        SR-RAG & \underline{60.0} & 72.6 & 38.8 & 54.2 \\
        FS-RAG & 59.5 & \underline{73.8} & 25.9 & 36.5 \\
        FLARE & 49.8 & 67.9 & 38.7 & 52.1 \\
        Web-Tool  & 42.9 & 63.2 & 8.9 & 16.8 \\
        \rowcolor{lightblue!20} QuCo-RAG & \textbf{64.6} & \textbf{74.8} & \textbf{48.2} & \textbf{62.2} \\
        
        \midrule
        \multicolumn{5}{l}{\textbf{GPT-5-chat}} \\
        \midrule
        Wo-RAG & 50.1 & 67.0 & 37.7 & 54.5 \\
        SR-RAG & \underline{51.0} & \underline{70.1} & \underline{42.9} & \underline{58.6} \\
        FS-RAG & 47.3 & 63.3 & 19.0 & 31.3 \\
        Web-Tool  & 48.3 & 69.8 & 19.8 & 33.6 \\
        \rowcolor{lightblue!20} QuCo-RAG & \textbf{59.7} & \textbf{73.3} & \textbf{48.4} & \textbf{62.6} \\
        \bottomrule
    \end{tabular}
    }
\end{table}

\subsection{Detailed Efficiency Metrics}
\label{sec:appendix_efficiency}

Table~\ref{tab:efficiency_merged} presents the complete efficiency comparison across all OLMo-2 model sizes on both datasets. We report three metrics: average token consumption (\#Tok.), LLM calls (\#Call), and retrieval operations (\#Ret.) per question. QuCo-RAG maintains competitive efficiency across all settings. Notably, on HotpotQA with OLMo-2-32B, QuCo-RAG achieves the highest EM (41.6, see Table~\ref{tab:main_results}) while using only 98 tokens and 1.90 LLM calls, compared to FS-RAG which consumes 594 tokens and 8.59 calls yet achieves only 13.9 EM. SeaKR consistently incurs the highest number of LLM calls (9--14 per question) due to its iterative hidden-state uncertainty estimation.

\begin{table}[h]
    \centering
    \caption{Efficiency comparison of RAG methods across OLMo-2 model sizes. \#Tok.: average number of tokens used; \#Call: average number of LLM calls; \#Ret.: average number of retrieval operations.}
    \label{tab:efficiency_merged}
    \begin{adjustbox}{max width=\columnwidth}
    \begin{tabular}{lcccccc}
        \toprule
        & \multicolumn{3}{c}{2WikiMultihopQA} & \multicolumn{3}{c}{HotpotQA} \\
        \cmidrule(lr){2-4} \cmidrule(lr){5-7}
        Method & \#Tok. & \#Call & \#Ret. & \#Tok. & \#Call & \#Ret. \\
        \midrule
        \multicolumn{7}{l}{\textbf{OLMo-2-7B}} \\
        \midrule
        Wo-RAG & 58.62 & 1.00 & 0.00 & 54.15 & 1.00 & 0.00 \\
        SR-RAG & 49.23 & 1.00 & 1.00 & 69.04 & 1.00 & 1.00 \\
        FS-RAG & 306.09 & 4.96 & 4.96 & 417.77 & 6.91 & 6.91 \\
        FLARE  & 132.90 & 2.33 & 1.03 & 436.37 & 6.89 & 3.39 \\
        DRAGIN & 114.09 & 2.58 & 1.27 & 387.54 & 6.52 & 3.24 \\
        ETC    & 124.48 & 3.25 & 1.25 & 83.69 & 2.38 & 0.79 \\
        SeaKR  & 99.89 & 11.91 & 1.39 & 100.22 & 10.95 & 1.29 \\
        \rowcolor{new_red!20}
        QuCo-RAG & 107.87 & 2.44 & 2.61 & 128.20 & 3.23 & 4.47 \\
        \midrule
        \multicolumn{7}{l}{\textbf{OLMo-2-13B}} \\
        \midrule
        Wo-RAG & 53.63 & 1.00 & 0.00 & 54.59 & 1.00 & 0.00 \\
        SR-RAG & 70.65 & 1.00 & 1.00 & 69.57 & 1.00 & 1.00 \\
        FS-RAG & 234.42 & 4.36 & 4.36 & 464.35 & 6.48 & 6.48 \\
        FLARE  & 129.67 & 2.01 & 0.93 & 284.34 & 3.42 & 1.69 \\
        DRAGIN & 134.78 & 2.78 & 1.27 & 254.14 & 4.26 & 1.96 \\
        ETC    & 126.00 & 3.23 & 1.22 & 100.26 & 2.56 & 0.85 \\
        SeaKR  & 78.42 & 9.42 & 1.01 & 92.11 & 10.28 & 1.29 \\
        \rowcolor{new_red!20}
        QuCo-RAG & 105.83 & 2.50 & 2.50 & 87.19 & 1.84 & 1.70 \\
        \midrule
        \multicolumn{7}{l}{\textbf{OLMo-2-32B}} \\
        \midrule
        Wo-RAG & 54.72 & 1.00 & 0.00 & 76.19 & 1.00 & 0.00 \\
        SR-RAG & 64.61 & 1.00 & 1.00 & 91.31 & 1.00 & 1.00 \\
        FS-RAG & 266.70 & 5.02 & 5.02 & 593.71 & 8.59 & 8.59 \\
        FLARE  & 116.19 & 2.10 & 1.01 & 270.10 & 3.20 & 1.59 \\
        DRAGIN & 103.53 & 2.69 & 1.26 & 554.09 & 7.49 & 3.71 \\
        ETC    & 116.85 & 3.15 & 1.19 & 106.24 & 2.61 & 0.91 \\
        SeaKR  & 91.08 & 14.26 & 2.46 & 79.43 & 12.72 & 1.97 \\
        \rowcolor{new_red!20}
        QuCo-RAG & 116.29 & 2.43 & 2.49 & 98.09 & 1.90 & 1.99 \\
        \bottomrule
    \end{tabular}
    \end{adjustbox}
\end{table}

\begin{figure}[h]
    \centering
    \includegraphics[width=0.98\linewidth]{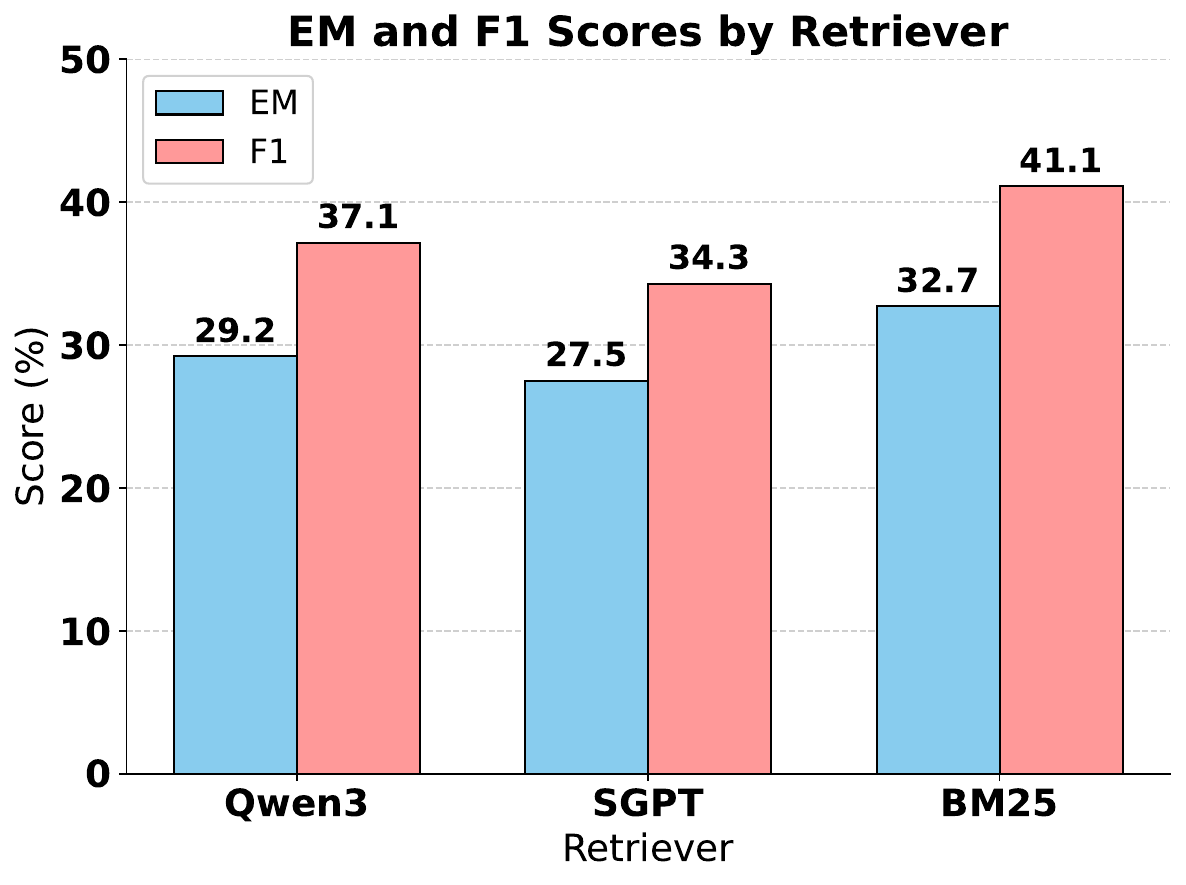}
    \caption{Performance comparison of QuCo-RAG with different retrievers (Qwen3-Embedding, SGPT, and BM25) on 2WikiMultihopQA using OLMo-2-7B.}
    \label{fig:retriever_comparison}
\end{figure}

\subsection{Effect of Different Retrievers}
\label{app:retriever_robustness}
To verify that QuCo-RAG is robust to retriever choice, we compare BM25 with dense retrievers SGPT~\cite{muennighoff2022sgpt} and Qwen3-Embedding-0.6B~\cite{zhang2025qwen3}.
As shown in Figure~\ref{fig:retriever_comparison}, QuCo-RAG achieves robust performance across all three retrievers, with EM scores ranging from 27.5 to 32.7 and F1 from 34.3 to 41.1. BM25 achieves the best results (32.7 EM, 41.1 F1), aligning with prior findings that sparse retrieval remains highly competitive for RAG tasks~\cite{su-etal-2024-dragin}. Importantly, even with different retriever backends, QuCo-RAG consistently outperforms baselines (cf. Table~\ref{tab:main_results}), confirming that our corpus-based uncertainty quantification mechanism is orthogonal to the choice of retrieval system.

\subsection{Sensitivity Analysis}
\label{sec:threshold_sensitivity}

We examine the robustness of QuCo-RAG to its key design choices: the entity frequency threshold $\tau_{\text{entity}}$, the co-occurrence threshold $\tau_{\text{cooc}}$, the co-occurrence window size $\omega$, and the Stage~1 aggregation strategy.\footnote{The ablation results in 
this section are from an independent evaluation run; the relative 
rankings are preserved. See Table~\ref{tab:main_results} for the main 
results.}

\begin{figure}[h]
    \centering
    \begin{minipage}[b]{0.48\textwidth}
        \centering
        \includegraphics[width=\textwidth]{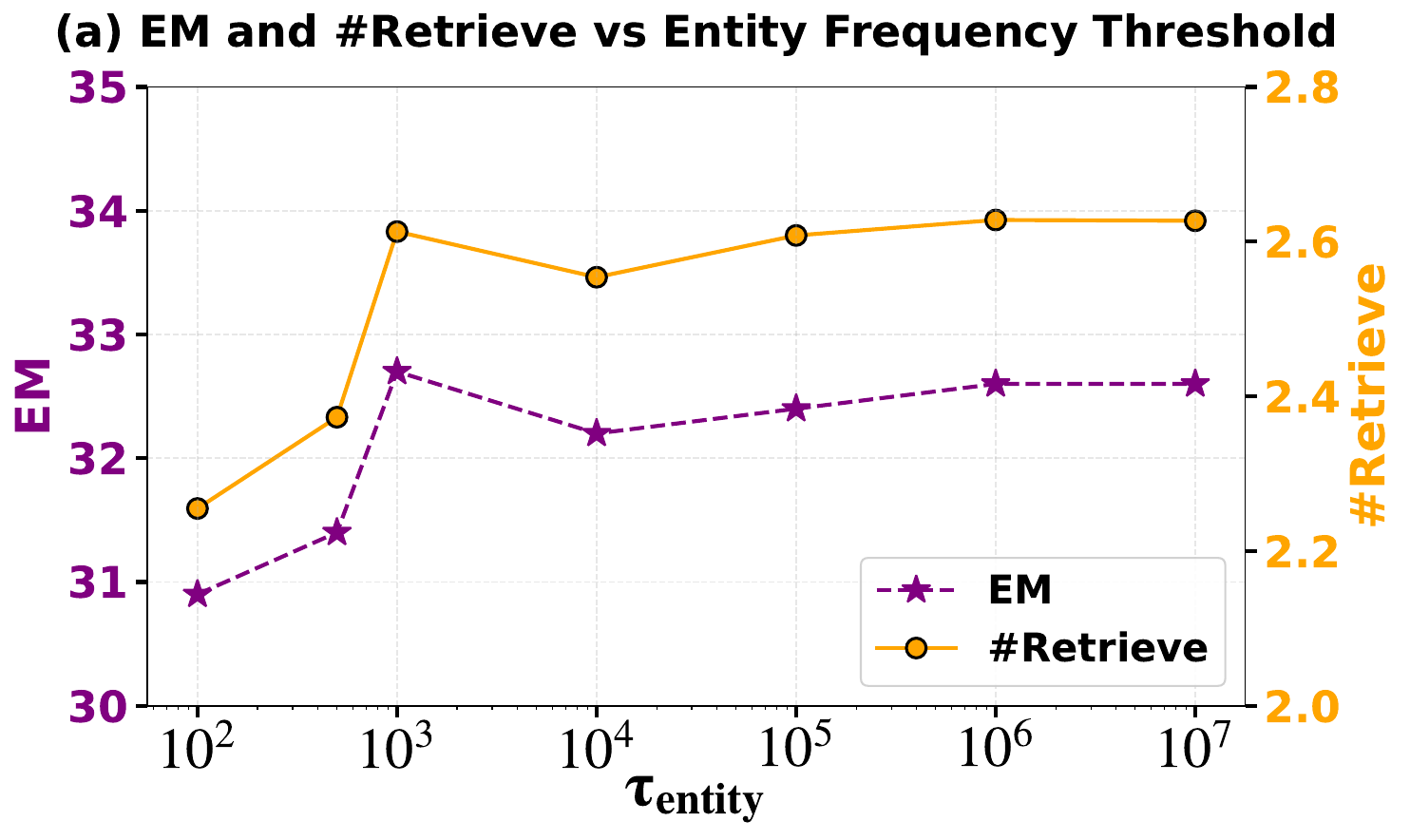}
    \end{minipage}%
    \hspace{0.5cm}%
    \begin{minipage}[b]{0.48\textwidth}
        \centering
        \includegraphics[width=\textwidth]{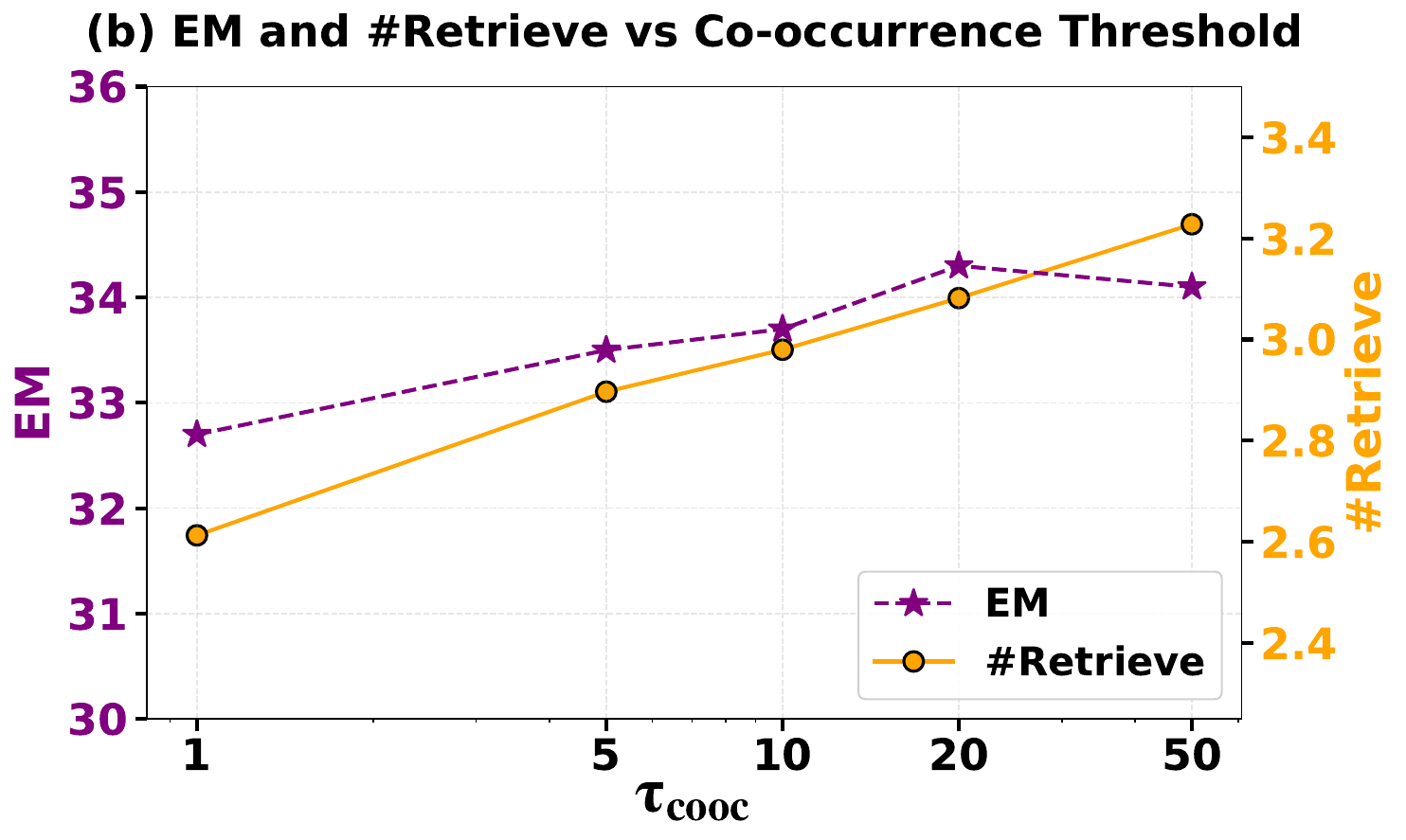}
    \end{minipage}
    \caption{Threshold sensitivity analysis on 2WikiMultihopQA with OLMo-2-7B.}
    \label{fig:threshold_sensitivity}
\end{figure}

\paragraph{Entity Frequency Threshold.}
As illustrated in Figure~\ref{fig:threshold_sensitivity}(a), EM remains stable (32.2--32.7) across a wide range of $\tau_{\text{entity}}$ from $10^3$ to $10^7$, with retrieval count also staying consistent (2.5--2.6), demonstrating strong robustness to this hyperparameter. 

\paragraph{Co-occurrence Threshold.}
As shown in Figure~\ref{fig:threshold_sensitivity}(b), increasing the threshold imposes a stricter verification standard (requiring more evidential support in the corpus), leading to a monotonic increase in retrieval frequency (from 2.61 to 3.23). While higher thresholds (e.g., $\tau_{cooc}=20$) yield marginal EM improvements (reaching 34.3 EM), they incur significantly higher retrieval overhead. 
We adopt $\tau_{cooc} = 1$ (i.e., triggering on zero co-occurrence) as our default for its clear interpretability: if two entities never co-occur in the pre-training corpus, the generated claim lacks evidential support and is likely hallucinated.

\paragraph{Co-occurrence Window Size.}
Table~\ref{tab:window_sensitivity} reports results across $\omega = 50$ to $2{,}000$ on both benchmarks.
EM varies by at most 1.4 points across the full range, demonstrating strong robustness.
Larger windows yield higher co-occurrence counts, reducing Stage~2 retrieval triggers, while smaller windows impose stricter locality constraints and trigger more retrievals.
Stage~1 retrieval remains constant across all settings (${\sim}0.75$ on 2Wiki, ${\sim}0.71$ on HotpotQA), since the window size only affects Stage~2.
We adopt $\omega = 1{,}000$ as the default because it roughly matches passage-level context length, providing a natural semantic boundary for co-occurrence verification.

\begin{table}[h]
\centering
\resizebox{\columnwidth}{!}{
\begin{tabular}{r cc cc}
\toprule
& \multicolumn{2}{c}{\textbf{2WikiMultihopQA}} & \multicolumn{2}{c}{\textbf{HotpotQA}} \\
\cmidrule(lr){2-3} \cmidrule(lr){4-5}
$\omega$ & EM & \#Ret.\,(S2) & EM & \#Ret.\,(S2) \\
\midrule
50   & 32.8 & 2.77\,(2.03) & 35.8 & 3.11\,(2.40) \\
100  & 32.1 & 2.66\,(1.92) & 35.7 & 2.95\,(2.24) \\
250  & 32.1 & 2.55\,(1.80) & 35.7 & 2.81\,(2.10) \\
500  & 31.7 & 2.48\,(1.74) & 35.1 & 2.75\,(2.04) \\
1000$^\dagger$ & 31.5 & 2.43\,(1.68) & 35.4 & 2.70\,(2.00) \\
2000 & 31.4 & 2.39\,(1.64) & 35.1 & 2.68\,(1.97) \\
\bottomrule
\end{tabular}
}
\caption{Co-occurrence window size ($\omega$) sensitivity on OLMo-2-7B. \#Ret.: total retrieval count per question; (S2): Stage~2 only. $^\dagger$: default.}
\label{tab:window_sensitivity}
\end{table}

\paragraph{Stage~1 Aggregation Strategy.}
We compare three strategies for aggregating entity frequencies in Stage~1: minimum, average, and maximum.
As shown in Table~\ref{tab:aggregation}, all three yield comparable EM (within 0.5 points), with Stage~2 retrieval counts remaining nearly constant across settings.
This robustness stems from the two-stage design: Stage~2 effectively compensates for cases that Stage~1 may miss.
We adopt average as the default because it provides a balanced measure of overall knowledge coverage for the input question.

\begin{table}[h]
\centering
\resizebox{\columnwidth}{!}{
\begin{tabular}{l cc cc}
\toprule
& \multicolumn{2}{c}{\textbf{2WikiMultihopQA}} & \multicolumn{2}{c}{\textbf{HotpotQA}} \\
\cmidrule(lr){2-3} \cmidrule(lr){4-5}
Agg. & EM & \#Ret.\,(S1/S2) & EM & \#Ret.\,(S1/S2) \\
\midrule
Min            & 31.6 & 2.52\,(0.84/1.68) & 35.2 & 2.73\,(0.76/1.98) \\
Avg$^\dagger$  & 31.5 & 2.43\,(0.75/1.68) & 35.4 & 2.70\,(0.71/1.99) \\
Max            & 31.3 & 2.41\,(0.72/1.69) & 35.7 & 2.69\,(0.68/2.00) \\
\bottomrule
\end{tabular}
}
\caption{Stage~1 aggregation strategy ablation on OLMo-2-7B. \#Ret.: total retrieval count; (S1/S2): Stage~1 and Stage~2 counts. $^\dagger$: default.}
\label{tab:aggregation}
\end{table}

\subsection{Detailed Performance Breakdown by Entity Frequency Bin}
\label{app:frequency_breakdown}

Table~\ref{tab:frequency_breakdown} presents the full performance breakdown by entity frequency. Entity frequency is defined as the average occurrence count of all entities in the question within the OLMo-2 pre-training corpus. QuCo-RAG achieves the best EM in 6 out of 8 frequency bins, with particularly large gains on low-frequency entities (frequency < 50) where internal-signal-based methods (FLARE, DRAGIN) perform similarly to Wo-RAG. This validates our core hypothesis that entity frequency in the pre-training corpus serves as an effective indicator of knowledge gaps.

\begin{table*}[t]
    \centering
    \caption{Detailed performance breakdown by entity frequency on 2WikiMultihopQA (OLMo-2-7B). Entity frequency is defined as the average appearance count of all entities in the question within the OLMo-2 pre-training corpus.}
    \label{tab:frequency_breakdown}
    \begin{adjustbox}{max width=0.94\textwidth}
    \begin{tabular}{lc|cc|cc|cc|cc|cc|cc}
        \toprule
        & & \multicolumn{2}{c|}{Wo-RAG} & \multicolumn{2}{c|}{SR-RAG} & \multicolumn{2}{c|}{FS-RAG} & \multicolumn{2}{c|}{FLARE} & \multicolumn{2}{c|}{DRAGIN} & \multicolumn{2}{c}{QuCo-RAG} \\
        Freq. Bin & Count & EM & \#Ret. & EM & \#Ret. & EM & \#Ret. & EM & \#Ret. & EM & \#Ret. & EM & \#Ret. \\
        \midrule
        0 & 180 & 12.8 & 0.00 & 13.9 & 1.00 & 14.4 & 4.52 & 11.1 & 0.97 & 12.2 & 1.26 & \textbf{22.8} & 2.25 \\
        1-10 & 117 & 11.1 & 0.00 & 20.5 & 1.00 & 15.4 & 4.62 & 13.7 & 0.87 & 13.7 & 1.31 & \textbf{28.2} & 2.41 \\
        11-50 & 119 & 13.4 & 0.00 & 25.2 & 1.00 & 18.5 & 4.79 & 17.6 & 0.84 & 15.1 & 1.32 & \textbf{26.9} & 2.67 \\
        51-100 & 66 & 27.3 & 0.00 & 18.2 & 1.00 & 16.7 & 5.15 & 25.8 & 1.17 & \textbf{36.4} & 1.18 & 34.8 & 2.91 \\
        101-500 & 198 & 23.2 & 0.00 & 21.2 & 1.00 & 23.7 & 4.94 & 28.3 & 0.97 & 23.7 & 1.29 & \textbf{32.8} & 2.76 \\
        501-1k & 71 & 29.6 & 0.00 & \textbf{40.8} & 1.00 & 29.6 & 5.13 & 33.8 & 0.89 & 35.2 & 1.24 & 39.4 & 2.90 \\
        1k-5k & 141 & 24.1 & 0.00 & 29.1 & 1.00 & 24.8 & 5.38 & 31.2 & 1.23 & 31.9 & 1.28 & \textbf{41.8} & 2.81 \\
        >5k & 108 & 25.9 & 0.00 & 29.6 & 1.00 & 27.8 & 5.53 & 27.8 & 1.37 & 29.6 & 1.25 & \textbf{42.6} & 2.48 \\
        \midrule
        Overall & 1000 & 19.9 & 0.00 & 23.5 & 1.00 & 21.0 & 4.96 & 22.8 & 1.03 & 22.9 & 1.27 & \textbf{32.7} & 2.61 \\
        \bottomrule
    \end{tabular}
    \end{adjustbox}
\end{table*}

\subsection{Analysis of Co-occurrence Verification}
\label{sec:cooc_analysis}
 
As discussed in \S\ref{sec:runtime_verification}, Stage~2 verifies entity co-occurrence rather than full relational claims, because relational predicates exhibit high lexical variability while named entities are more lexically stable.
This design is intentionally asymmetric: $\text{cooc}(h,t) = 0$ strongly signals hallucination risk, but $\text{cooc}(h,t) > 0$ does not guarantee relational correctness---the entities may co-occur under different relations or in unrelated contexts.
We therefore conduct a post-hoc analysis to quantify how this asymmetry affects Stage~2's ability to detect wrong-relation hallucinations.

\paragraph{False Negative Analysis.}
We randomly sample 200 incorrect predictions (EM=0) per dataset and use GPT-5.2-Thinking for sentence-level hallucination annotation, yielding 183 and 202 annotated hallucinated sentences on HotpotQA and 2Wiki respectively.
For each hallucinated sentence, we extract triplets with our extractor and check whether Stage~2 flagged it.
The goal is not to estimate overall system accuracy, but to locate Stage~2's primary failure mode.
Results are shown in Table~\ref{tab:false_negative}.

\begin{table}[h]
\centering
\resizebox{\columnwidth}{!}{
\begin{tabular}{lcc}
\toprule
& \textbf{HotpotQA} & \textbf{2Wiki} \\
\midrule
Hallucinated sent.             & 183  & 202  \\
Detected by S2 (TP)            & 95\,(51.9\%) & 95\,(47.0\%) \\
Missed by S2 (FN)              & 88\,(48.1\%) & 107\,(53.0\%) \\
\quad Wrong relation in FN     & 60\,(68.2\%) & 63\,(58.9\%) \\
\midrule
Detection w/ + Rel.\ Check     & 149/183\,(81.4\%) & 156/202\,(77.2\%) \\
\bottomrule
\end{tabular}
}
\caption{Post-hoc false negative analysis on 200 incorrect predictions per dataset (OLMo-2-7B). S2: Stage~2; TP: hallucination detected; FN: hallucination missed. The last row shows detection rate after adding the relation-aware extension.}
\label{tab:false_negative}
\end{table}

Among the incorrectly answered questions, Stage~2 detects roughly half of the hallucinated sentences.
Of the missed cases, approximately 59--68\% involve wrong-relation hallucinations: the head and tail entities genuinely co-occur in the corpus but the generated relation is incorrect.
This suggests that wrong-relation errors account for a notable portion of Stage~2's missed detections.

\paragraph{Relation-Aware Extension.}
Motivated by this analysis, we test a simple optional extension: when $\text{cooc}(h,t) > 0$ (i.e., Stage~2 would not trigger retrieval), we additionally query the n-gram frequency of the concatenated phrase ``$h$ + $r$ + $t$'' in the corpus.
If this frequency is zero---indicating that the specific relational claim has no evidential support despite entity co-occurrence---we trigger retrieval.
As shown in Table~\ref{tab:relation_aware}, this extension raises the hallucination detection rate from ${\sim}50\%$ to ${\sim}77$--$81\%$ (Table~\ref{tab:false_negative}, last row), yielding $+2.1$ to $+2.4$ EM improvement end-to-end.
 
\begin{table}[h]
\centering
\resizebox{\columnwidth}{!}{
\begin{tabular}{lcccccc}
\toprule
& \multicolumn{3}{c}{\textbf{2WikiMultihopQA}} & \multicolumn{3}{c}{\textbf{HotpotQA}} \\
\cmidrule(lr){2-4} \cmidrule(lr){5-7}
Method & EM & F1 & \#Ret. & EM & F1 & \#Ret. \\
\midrule
QuCo-RAG             & 31.5 & 39.8 & 2.43 & 35.4 & 46.5 & 2.70 \\
\quad + Rel.\ Check  & 33.6 & 41.6 & 3.68 & 37.8 & 47.9 & 3.74 \\
\midrule
$\Delta$             & \textcolor{darkred}{\textbf{+2.1}} & +1.8 & +1.25 & \textcolor{darkred}{\textbf{+2.4}} & +1.4 & +1.04 \\
\bottomrule
\end{tabular}
}
\caption{End-to-end ablation of the relation-aware extension (OLMo-2-7B).}
\label{tab:relation_aware}
\end{table}

This improvement comes at the cost of increased retrieval frequency (about 39--51\% in total retrievals), which is why we retain the original co-occurrence check as the default configuration and present the relation-aware variant as an optional enhancement when accuracy is prioritized over latency.
Future work may further close this gap through relation canonicalization, entailment-based verification, or integrating semantic matching with entity linking.

\end{document}